\documentclass[runningheads]{llncs}
\PassOptionsToPackage{table}{xcolor}

\usepackage[normalem]{ulem}
\usepackage{eccv}
\usepackage{array}
\usepackage{graphicx}
\usepackage{amsmath}
\usepackage{amssymb}
\usepackage{booktabs}
\usepackage{bbold}
\usepackage{multirow}
\usepackage{bm}
\usepackage{arydshln}
\usepackage{xspace}
\usepackage{url}
\usepackage{comment}
\usepackage{mathabx}
\usepackage{wasysym}
\usepackage{capt-of}
\usepackage{adjustbox}
\usepackage[dvipsnames]{xcolor}
\usepackage{times}
\usepackage{epsfig}
\usepackage{amssymb}
\usepackage{cuted}
\usepackage{algorithm}
\usepackage{algpseudocode}

\definecolor{GreenColor}{rgb}{0.137,0.573,0.565}
\definecolor{OrangeColor}{rgb}{0.914,0.541,0.0.141}
\definecolor{PurpleColor}{rgb}{0.5,0,0.7}
\definecolor{BlueColor}{rgb}{0,0.725,0.949}
\definecolor{PinkColor}{rgb}{0.9843,0.19215,0.6}

\makeatletter
\newcommand{\figcaption}[1]{\def\@captype{figure}\caption{#1}}
\newcommand{\tblcaption}[1]{\def\@captype{table}\caption{#1}}
\newcommand{\customparagraph}[1]{\par{\noindent\textbf{#1:}}}
\newcommand{\textcite}[1]{``\textit{#1}''}

\makeatother

\def\chi{Proceedings of the SIGCHI Conference on Human Factors in Computing Systems (CHI)}

\makeatletter
\DeclareRobustCommand\onedot{\futurelet\@let@token\@onedot}
\def\@onedot{\ifx\@let@token.\else.\null\fi\xspace}
\def\eg{\emph{e.g}\onedot} 
\def\ie{\emph{i.e}\onedot} 
 
 \def\vs{\emph{vs}\onedot}

\makeatother

\newcommand{\vdashrule}{\vrule width 0.8pt}
\usepackage{caption}
\usepackage{subcaption}
\usepackage{algorithm}
\usepackage{algpseudocode}
\usepackage{pgfplots}
\usepackage{pgfplotstable}
\pgfplotsset{compat=1.18}
\pgfplotsset{
    colormap={softgreen}{
        rgb255=(255, 255, 255) 
        rgb255=(225, 245, 225) 
        rgb255=(180, 230, 180) 
        rgb255=(110, 190, 110)  
    }
}

\providecommand{\samethanks}[1][\value{footnote}]{\footnotemark[#1]}

\definecolor{dark_green}{rgb}{0, 0.5, 0}

\usepackage{eccvabbrv}

\usepackage{graphicx}
\usepackage{booktabs}

\usepackage[accsupp]{axessibility}

\usepackage[breaklinks,hidelinks]{hyperref}

\usepackage{orcidlink}
\usepackage{arydshln}
\usepackage[table]{xcolor}

\renewcommand{\customparagraph}[1]{\par\noindent\textbf{#1:}}

\begin{document}

\title{Affordance-Guided Diffusion Prior for\\3D Hand Reconstruction}

\titlerunning{Affordance-Guided Diffusion Prior}

\author{
Naru Suzuki\inst{1}\thanks{Equal contribution.} \and
Takehiko Ohkawa\inst{1}\samethanks \and
Tatsuro Banno\inst{1} \and\\
Jihyun Lee\inst{2} \and
Ryosuke Furuta\inst{1} \and
Yoichi Sato\inst{1}
}
\authorrunning{N.~Suzuki et al.}
\institute{%
$^{1}$The University of Tokyo \qquad
$^{2}$KAIST
}

\maketitle

\begin{abstract}
How can we infer a 3D hand pose when large portions of the hand are heavily occluded by itself or by objects?
Humans often resolve such ambiguities by leveraging contextual knowledge—such as \emph{affordances}, where an object’s shape and function suggest how the object is typically grasped.
Inspired by this observation, we propose a generative prior for 3D hand pose modeling guided by affordance-aware textual descriptions of hand-object interactions (HOI).
Our method employs a diffusion-based generative model that learns the distribution of plausible hand poses conditioned on 
contextual signals, such as affordance descriptions and image features. 
The affordance descriptions are designed to represent the semantic intent and geometric structure of HOI, using the reasoning from a vision-language model (VLM) and grasp classification. 
We leverage the diffusion prior to refine the 3D pose predictions in hand reconstruction into more accurate and functionally coherent estimation.
Our experiments demonstrate that our affordance-guided refinement significantly improves 3D hand pose estimation performance on 3D hand affordance datasets, HOGraspNet and HO3D,
over state-of-the-art methods such as 
foundation models for hand reconstruction and the latest diffusion priors for 3D hands.
\end{abstract}

\section{Introduction}
\label{sec:intro}
Hand-object interactions (HOI) are ubiquitous in daily life, where human actions are inseparable from the surrounding context, including objects, environments, and the person's intent.
Among them, reconstructing 3D hands (\eg, estimating pose and shape parameters of the MANO model~\cite{Romero:tog17:mano}) from a single RGB image has become a crucial task, enabling applications in virtual/augmented reality and robotic dexterous manipulation.
However, this task remains highly challenging due to severe occlusions and inherent 3D ambiguities commonly present in real-life interactions.
Despite these difficulties, humans can still infer plausible hand poses by exploiting \emph{contextual knowledge} beyond the hand itself.
In particular, the shape and function of interacting objects can provide strong cues for reasoning about feasible hand configurations during interaction.

\begin{figure}[t]
    \hfill
    \includegraphics[width=\linewidth]{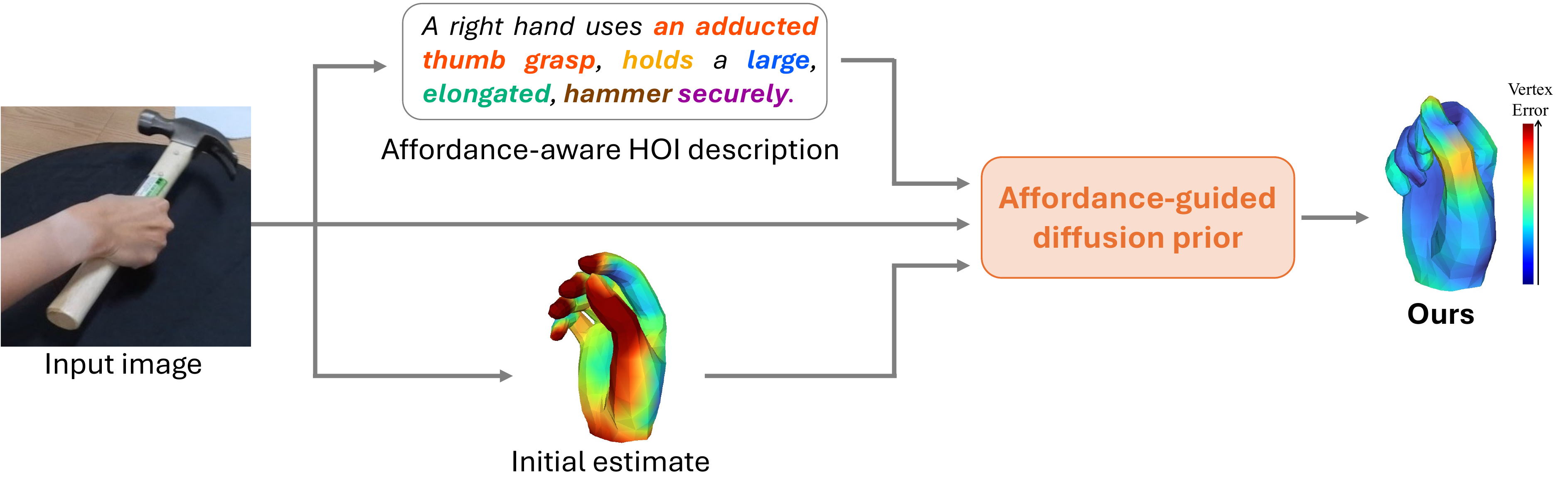}
    \vspace{-7mm}
    \caption{
    \textbf{Can affordance-aware semantics benefit 3D hand reconstruction?}
    By leveraging contextual cues about how an object is grasped, our affordance-guided diffusion prior refines coarse 3D hand predictions into functionally coherent and physically grounded poses, even under strong occlusion. Vertex error is color-coded on the hand mesh.
    }
    \label{fig:teaser_1}
\end{figure}

Most previous works on 3D hand reconstruction~\cite{pavlakos:cvpr24:HaMeR,potamias:cvpr25:wilor,fan:eccv24:benchmarks,park:cvpr22:handoccnet} have paid little attention to leveraging such contextual knowledge.
Only a few studies have explored contextual signals in the following forms: (a) temporal contact states (\eg, 
pre-grasp, in-grasp, post-grasp)~\cite{zhang:iccv25:intuitivephys}, (b) short interaction captions (\eg, ``a hand holding a spoon'')~\cite{Ye:cvpr24:G-HOP}, (c) action labels (\eg, ``screw'', ``open'')~\cite{ohkawa:cvpr23:assemblyhands,liu:cvpr22:hoi4d}, (d) intention labels (\eg, intend to ``use'' or ``receive'' an object)~\cite{taheri:eccv20:grab,yang:cvpr22:oakink}, etc.
These semantic labels have been used as conditional input in 3D modeling~\cite{wang:cvpr25:promptHMR,Ye:cvpr24:G-HOP,yang:cvpr22:oakink}, for additional semantic classification~\cite{prakash:eccv24:everydayego,wen:cvpr23:htt,zhang:iccv25:intuitivephys}, or to create specific evaluation categories~\cite{liu:cvpr22:hoi4d,ohkawa:cvpr23:assemblyhands,cho:eccv24dense}.
However, these existing labels remain coarse and fail to capture the rich details of HOI needed to resolve ambiguities. 

To enrich contextual signals and better reflect the subtle nuances of HOI scenes, we emphasize the critical role of \emph{affordances}~\cite{corona:cvpr20:ganhand,goyal:cvpr22:human,bahl:cvpr23:affordances} as semantic guidance, which directly encodes how object properties constrain and guide possible hand interactions.
Existing affordance-related studies are typically categorized into recognizing (i) motor actions or (ii) grasp types, defined by discrete labels~\cite{yu:wacv23:affordanceego}.
Motor actions, often expressed by verb labels (\eg, ``pick up''), represent the high-level semantic intent of manipulation. 
In contrast, grasp types, defined through rigorous taxonomies (\eg, ``lateral grasp'', ``precision grasp'')~\cite{feix:2015:grasptaxonomy,cho:eccv24dense}, provide the geometric structure of fine-grained hand configurations.

Based on this perspective, we integrate these two distinct aspects of affordances, \ie, both semantic intent and geometric structure of HOI, into a unified contextual representation for hand reconstruction.
However, the way to combine these heterogeneous elements of affordances in a unified form is not obvious.
Unlike following existing categories (\eg, ``pick up'' or ``lateral grasp''), we tackle this issue by representing affordances through detailed textual descriptions that encode both semantic and geometric cues.
This descriptive form offers greater flexibility to capture variations in object attributes, hand grasps, and interaction intent beyond predefined categories.
In particular, we incorporate grasp types and object attributes (category, shape, and size), along with the underlying intent, such as interaction (action) and intention captions.
To generate such a descriptive text reflecting multiple affordance aspects, we propose a step-by-step generation process by leveraging visual commonsense reasoning of a large vision-language model (VLM)~\cite{bai2025qwen2.5vl} with grasp taxonomy knowledge from the hand affordance dataset~\cite{cho:eccv24dense}, resulting in complete and interpretable descriptions of affordances.

With the affordance descriptions, we propose generative prior modeling that can integrate textual descriptions into 3D pose modeling (Fig.~\ref{fig:teaser_1}).
We train a denoising diffusion model~\cite{ho:neurips20:ddpm} on hand pose data, while newly adopting affordance descriptions, encoded with DistilBERT~\cite{sanh2019distilbert}, and ViT-based image features~\cite{dosovitskiy:iclr21:vit} as conditional input.
We further utilize this prior to refine 3D pose estimates, inspired by~\cite{Ohkawa:iccv25:SCGen}.
Specifically, starting from an initial estimate by WiLoR~\cite{potamias:cvpr25:wilor} or Hamba~\cite{dong:neurips24:hamba}, we iteratively denoise the pose so that it remains consistent with the affordance-aware diffusion prior and visible 2D evidence.
To improve the refinement, we further detect occluded joints with ray casting from joints and hand segmentation to guide only uncertain joints.

Our experiments are conducted on 3D hand affordance datasets, HOGraspNet~\cite{cho:eccv24dense} and HO3D~\cite{hampali:cvpr20:ho3d} where the grasp type labels are assigned by the classification model trained on HOGraspNet. 
Our affordance-guided diffusion prior consistently outperforms recent hand reconstruction models, including foundational transformers such as WiLoR~\cite{potamias:cvpr25:wilor} and HaMeR~\cite{pavlakos:cvpr24:HaMeR}, the Mamba-based Hamba~\cite{dong:neurips24:hamba}, and hand diffusion priors based on InterHandGen~\cite{lee:cvpr24:interhandgen}, with or without contextual reasoning.
Furthermore, our qualitative study finds our method producing grasp poses that not only appear more natural but also align closely with the affordance descriptions.

Our main contributions are summarized as follows:
\begin{itemize}
\vspace{-3mm}
\item We are the first to introduce affordance-aware textual descriptions into 3D hand reconstruction using VLMs and grasp classification.
\item We design a diffusion-based generative prior that learns the hand pose distribution given the contextual signals, enabling semantically consistent generation.
\item We propose an occlusion-aware refinement approach that leverages the affordance-aware diffusion prior for 3D hand reconstruction, yielding substantial improvements in pose accuracy specifically when estimated joints are uncertain.
\end{itemize}

\section{Related Work}
\customparagraph{3D hand reconstruction}
This task involves estimating hand pose and shape to reconstruct hand mesh, typically from a single RGB image.
The earlier efforts~\cite{park:cvpr22:handoccnet, kulon:cvpr20:weakly, zhang:iccv19:end, fan:eccv24:benchmarks, ohkawa:ijcv23:survey} encode image features with CNN-based backbones and regress pose and shape parameters of the MANO model~\cite{Romero:tog17:mano}.
Particularly, HandOccNet~\cite{park:cvpr22:handoccnet} proposes robust feature extraction for occluded regions. 
Recent foundation models~\cite{pavlakos:cvpr24:HaMeR,potamias:cvpr25:wilor} leverage a vision transformer~\cite{dosovitskiy:iclr21:vit} with large-scale pretraining to enhance generalization ability across different domains. 
While these recent methods typically provide strong alignment to 2D observation, ambiguities still remain in the 3D space, \eg, depth ambiguity and uncertainty for occluded joints.

To address this, integrating with contextual cues is crucial as humans can infer feasible 3D poses given the context.
Yet, this direction remains relatively underexplored.
For instance, recent methods~\cite{prakash:eccv24:everydayego,wang:cvpr25:promptHMR,zhang:iccv25:intuitivephys,wen:cvpr23:htt,zhou:wacv26:dfmamba} attempt to integrate semantic or physical cues (\eg, grasp type and state, body proportion, and action labels), but focus on coarse-level reasoning rather than fine-grained details of HOI scenes.

\begin{figure}[t]
    \centering
    \includegraphics[width=\linewidth]{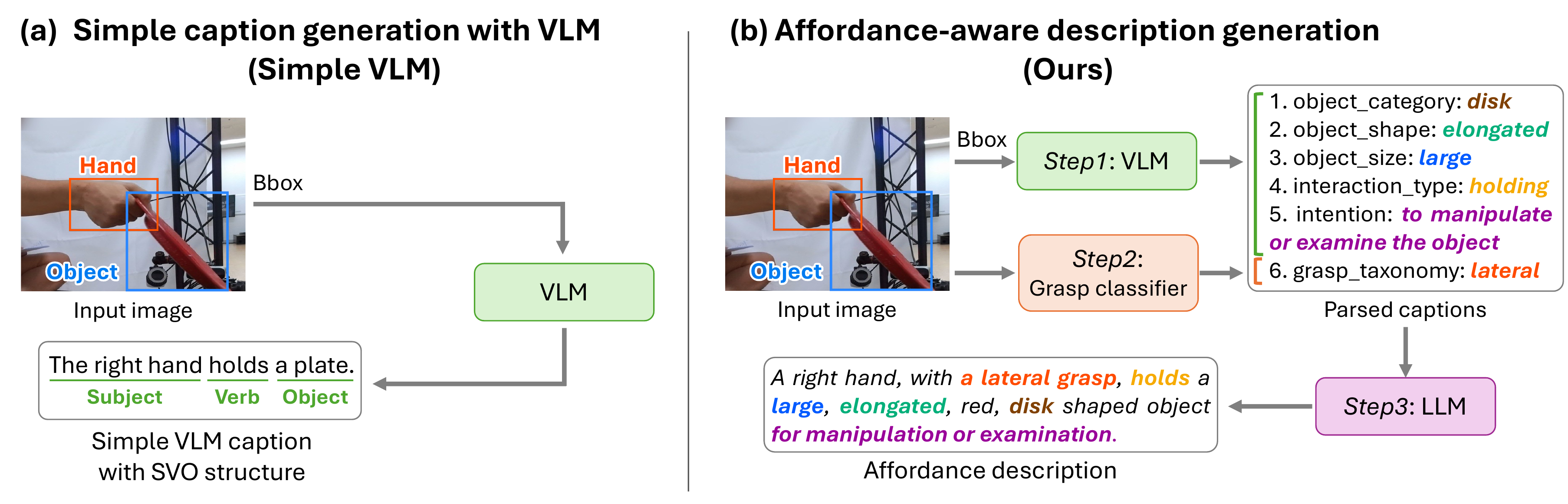}
    \caption{
    \textbf{Affordance description generation.} 
    (a) We show a naive baseline of caption generation (Simple VLM) from off-the-shelf VLMs  (\eg, Qwen2.5-VL~\cite{bai2025qwen2.5vl}) given hand-object bounding boxes, lacking fine-grained affordance cues. 
    (b) Inspired by chain-of-thought reasoning, our proposed generation first obtains parsed captions (items 1-5) by prompting the VLM to extract intended affordance elements. It further infers a grasp taxonomy (item 6) with a tailored classifier due to VLM's limited capability to predict hand grasps in a zero-shot manner.
    These generated affordance elements are summarized using LLM~\cite{albert:corr23:mistral7b}
    , resulting in a single detailed description.
    }
    \label{fig:aff_generation_scheme}
\end{figure}

In contrast, our work is motivated by leveraging detailed textual descriptions regarding affordances (see~\cref{fig:aff_generation_scheme}) and we are the first to introduce such affordance descriptions into 3D hand reconstruction.

\customparagraph{Affordance recognition and reasoning}
The concept of \emph{affordances}, introduced by Gibson~\cite{gibson2014theory}, describes the functional relationship between perception and potential action (\eg, a handle \emph{affords} grasping or a button \emph{affords} pressing). 
In computer vision, affordance recognition and reasoning have been studied through diverse formulations, including functional region segmentation~\cite{fang:cvpr18:demo2vec,jian:iccv23:affordpose,nagarajan:iccv19:grounded,li:cvpr24:one}, grasp and contact prediction~\cite{corona:cvpr20:ganhand,goyal:cvpr22:human,cho:eccv24dense}, affordance-based HOI understanding~\cite{koppula:13:learning,do:icra18:affordancenet} and generation~\cite{Cha:cvpr24:text2hoi,Ye:cvpr23:affordancediffusion,Zhou:nips25:megohand,Prakash:cvpr25:howdoIdothat},
and adaptation to robot manipulation~\cite{bahl:cvpr23:affordances,li:iccv25:affgrasp,wei:neurips24:graspasyousay}.
These approaches often operate at the object or scene level, capturing where and how an agent may interact.
Yet, they typically rely on coarse labels~\cite{yu:wacv23:affordanceego} such as verb categories (\eg, ``pick up'', ``place'') or predefined grasp taxonomies~\cite{feix:2015:grasptaxonomy}. 
Recent 3D hand affordance datasets~\cite{jian:iccv23:affordpose,cho:eccv24dense} further encourage affordance understanding with precise hand poses, object geometry, and vertex-wise contact.

In this work, we propose to represent affordances as rich, detailed, and multi-level semantic descriptions that unify motor intent, grasp configuration, and object attributes (see \cref{fig:aff_generation_scheme}). 
Furthermore, 
unlike the existing formulations,
we demonstrate that such affordance descriptions resolve the fine-grained geometric ambiguities central to 3D hand reconstruction.

\customparagraph{Diffusion models}
Denoising diffusion models~\cite{ho:neurips20:ddpm} have emerged as a powerful generative framework, owing to their strong capability to capture complex data distribution and their flexibility during inference.
Diffusion models are trained to gradually transform Gaussian noise into realistic samples through an iterative denoising process. 

Recent works adopt diffusion-based modeling in 3D hand and body modeling, outperforming the conventional VAEs-based modeling~\cite{lu:iccv25:dposer,muller:cvpr:24BUDDI}.
The diffusion models can leverage various conditional inputs for 3D modeling~\cite{Tevet:iclr23:MDM,christen:siga24:diffh2o, lee:cvpr24:interhandgen,Ohkawa:iccv25:SCGen,Ye:cvpr24:G-HOP,stathopoulos:cvpr:24scorehmr,liu:iccv25:tohgs}. 
Human motion generation commonly leverages text prompts to guide motion patterns~\cite{lv:eccv24:himo,Tevet:iclr23:MDM,tevet:eccv22:motionclip}, including hand-object motion generation~\cite{christen:siga24:diffh2o,li:cvpr25:latenthoi}.
InterHandGen~\cite{lee:cvpr24:interhandgen} synthesizes two-hand interactions conditioned on a single hand, while G-HOP~\cite{Ye:cvpr24:G-HOP} generates 3D hand-object interactions from object geometry.
The trained diffusion priors can be further adapted to 3D reconstruction tasks with score-distillation sampling~\cite{lee:cvpr24:interhandgen,Ye:cvpr24:G-HOP} and refinement with keypoint guidance~\cite{Ohkawa:iccv25:SCGen,stathopoulos:cvpr:24scorehmr}.

Our work instead proposes a pose diffusion model that flexibly takes multi-modal input for detailed text and visual features, unlike existing frameworks, \eg, text-to-motion~\cite{christen:siga24:diffh2o,li:cvpr25:latenthoi}, 3D two-hand~\cite{lee:cvpr24:interhandgen}, and 3D hand-object generation~\cite{Ye:cvpr24:G-HOP}.
It further enhances the diffusion-based refinement with guidance from occlusion detection unlike~\cite{Ohkawa:iccv25:SCGen}.

\section{Method}
Our work is motivated to leverage contextual signals from textual descriptions of affordances and visual evidence for 3D hand reconstruction.
To incorporate these cues, we propose a generative prior model based on a diffusion process that learns the 3D hand pose distribution given the contextual signals. 
Our approach begins with an affordance description generation, which extracts key elements of affordances from input images using a vision-language model and grasp classification, as described in \cref{sec:desc_gen}. 
We then construct a diffusion prior conditioned on the affordance descriptions and image features, 
as detailed in \cref{sec:description-conditioned}. 
Subsequently, the learned prior is adapted to correct occluded or uncertain joints with the refinement scheme in \cref{sec:3d_recon}, yielding more accurate and functionally coherent hand poses.

\subsection{Affordance description generation}\label{sec:desc_gen}
\emph{Affordances} represent how an object can be interacted with hands by capturing its actionable properties; thus we utilize them as contextual representation and guidance for 3D hand reconstruction.
Specifically, beyond standard categorical labels (\eg, grasp or action labels) of affordance recognition~\cite{cho:eccv24dense,koppula:13:learning,yu:wacv23:affordanceego,prakash:eccv24:everydayego}, we enrich its form with textual descriptions that reflect both semantic and precise geometric properties, such as object category, shape, size, interaction (action), intention, and grasp type. 

Our preliminary study finds that direct (one-step) captioning, which prompts VLMs to generate a description from an image, often produces a simplified sentence lacking fine-grained HOI properties.
An example of this simply generated caption (``Simple VLM'') is found in \cref{fig:aff_generation_scheme}(a).
Additionally, while being able to reasonably describe HOI scenes, recent VLMs still struggle to capture precise grasp taxonomy as this annotation is not common in general QA tasks or VLMs' pretraining data.
To provide finer affordance details, we generate the affordance descriptions step-by-step in \cref{fig:aff_generation_scheme}(b).
Inspired by the chain-of-thought reasoning~\cite{wei:neurips22:CoT} of LLMs, we break down the complex description task into small executable and interpretable steps. 
Our pipeline consists of three steps: (i) extracting key elements of affordances with zero-shot VLM reasoning, 
(ii) classifying the grasp with separate supervised training,
and (iii) integrating all elements into a final description with LLM. 
This procedure ensures the completeness and interpretability of the generated affordance descriptions with improved caption quality. We detail the individual steps below.

\customparagraph{Step 1 -- Extract parsed captions with VLM} We leverage a vision-language model (Qwen2.5-VL~\cite{bai2025qwen2.5vl}) to extract structured captions describing different aspects of affordances from an image. Specifically, given the image together with the bounding boxes of the hand and the object, we design a prompt that queries the model to describe five key elements that characterize HOI properties.
These include object category (\eg, ``disk'', ``sphere''), object shape (\eg, ``elongated'', ``flat''), object size (\eg, ``small'', ``large''), interaction (action) type (\eg, ``holding'', ``reaching''), 
and intention caption (\eg, ``to lift'', ``to manipulate'').
These elements are chosen because they capture the physical properties of the object and the semantic intent of the action.
Further details are found in the supplement.

\customparagraph{Step 2 -- Grasp classification}
Accessing grasp information is essential in determining hand configurations.
Yet, we observe that classifying grasp taxonomies using existing VLMs is highly challenging, as rigorous grasp labels~\cite{feix:2015:grasptaxonomy} may not be widely represented in their pretraining data.
Therefore, we train a dedicated grasp classification model independently on a large-scale hand affordance dataset, HOGraspNet~\cite{cho:eccv24dense}.
Specifically, we take the RGB image as input and extract visual features 
encoded by the ViT and MANO pose parameters using HaMeR~\cite{pavlakos:cvpr24:HaMeR}.
The visual and pose features are concatenated and passed through a self-attention layer, followed by an MLP layer for final classification.
Our model achieves 87.6\% accuracy in the evaluation dataset, and further implementation details and results are provided in the supplement.

\customparagraph{Step 3 -- Summarize with LLM} 
Given the short captions for each element related to affordances, we transform these items into a single detailed description. 
We use a large language model (LLM), Mistral-7B~\cite{albert:corr23:mistral7b}, with a summarization prompt that generates a final affordance description.
This LLM-based summarization allows us to combine nuanced variations in HOI properties, producing a coherent sentence that flexibly represents semantic and geometric cues beyond what predefined categories can express.

Overall, this divide-and-conquer approach comprised of the three steps ensures that the resulting descriptions are (i) complete and specific by eliciting all key elements, (ii) consistent with the predicted grasp type via supervised classification, and (iii) coherent and interpretable as the conditional input (used for the subsequent diffusion training).

\begin{figure}[t]
    \centering
    \includegraphics[width=\linewidth]{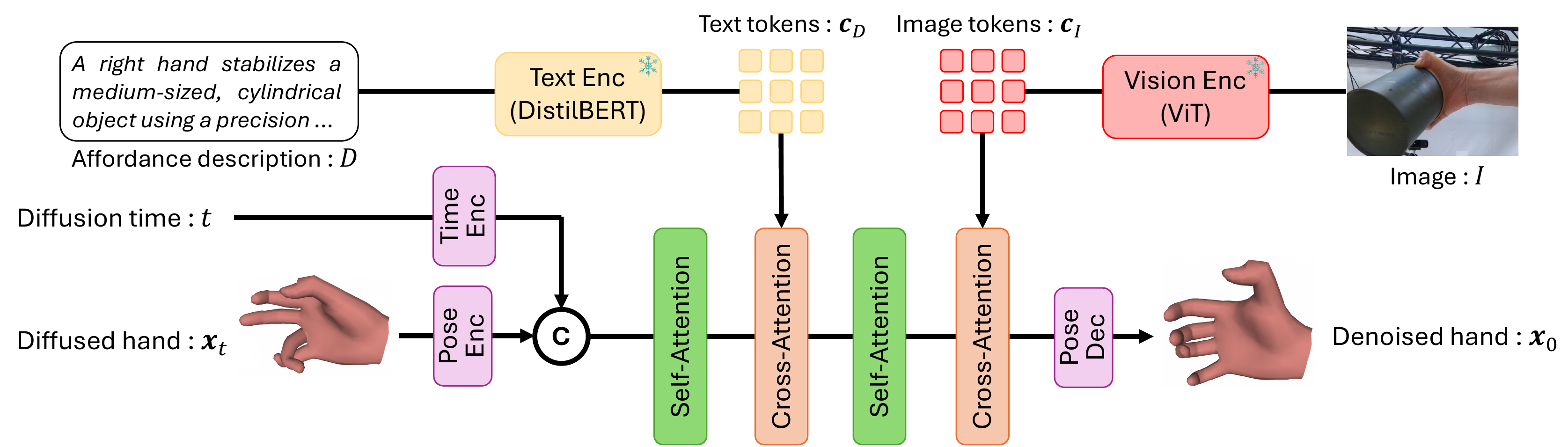}
    \caption{
    \textbf{Denoising framework of affordance-guided diffusion prior (AffHandGen).}  
    The input diffused hand pose is denoised using a transformer decoder network conditioned on the affordance description and RGB image, each encoded by DistilBERT~\cite{sanh2019distilbert} and the ViT backbone~\cite{pavlakos:cvpr24:HaMeR}. These modality interactions across pose, text, and image are learned through cross-attention.}
    \label{fig:arch_proposed}
\end{figure}

\subsection{Affordance-guided diffusion prior}\label{sec:description-conditioned}
Using the affordance descriptions, we present a diffusion prior model conditioned on text and image features, dubbed \textbf{AffHandGen}.
This model is trained to generate and refine 3D hand poses aligned with the contextual signals.
We choose the diffusion modeling to learn data distribution across different modalities, such as image, text, and pose, inspired by its successful results in 3D human synthesis~\cite{Tevet:iclr23:MDM,lee:cvpr24:interhandgen,patel:iccv25:uniegomotion,Ohkawa:iccv25:SCGen}.
It also benefits from being directly applicable to various downstream tasks without additional training~\cite{stathopoulos:cvpr:24scorehmr}, making it flexible and broadly reusable.

\customparagraph{Overview} 
With the goal of refining 3D hand pose in the downstream task, we construct a diffusion model in the hand pose space, where the shape estimation remains fixed. 
For the input, we use the pose parameters of MANO~\cite{Romero:tog17:mano} as the target variable of denoising, \ie, $\mathbf{x}_0=\boldsymbol{\theta}\in\mathbb{R}^{15\times 3}$.
We also use a diffusion timestep $t$, an affordance description $D$ (generated in~\cref{sec:desc_gen}), and the corresponding RGB image $I$ as the conditional input.
We encode the affordance description using a text encoder, yielding textual features $\mathbf{c}_D$, and encode the RGB image with an image encoder, yielding image features $\mathbf{c}_I$. 
The resulting diffusion model parameterized by $\phi$ is formulated as 
$\hat{\mathbf{x}}_0=f_{\phi}(\mathbf{x}_t, t, \mathbf{c})$, where $\mathbf{x}_t$ 
is the diffused pose at the timestep $t$ and $\mathbf{c}=\{\mathbf{c}_D,\mathbf{c}_I\}$ denotes the text and image condition.

The overall architecture of $f_{\phi}(\cdot)$ is illustrated in Fig.~\ref{fig:arch_proposed}.
Our \textbf{AffHandGen} is a transformer network trained to denoise noisy hand pose input $\mathbf{x}_t$ at the timestep $t$. 
The hand pose and timestep are first encoded through linear layers and concatenated, followed by processing with transformer decoder layers and a final pose decoder with a linear layer.

\customparagraph{Cross-attention to modality tokens}
For conditional data encoding, InterHandGen~\cite{lee:cvpr24:interhandgen} treats the condition $\mathbf{c}$ as a global feature vector and learns the relationships across the target pose, conditions, and timestep using self-attention. 
However, representing each conditional modality with a single vector loses spatial and semantic granularity.
For instance, when two hand images involve the same object but different grasp types, the global visual or textual embeddings (\eg, from CNN or CLIP~\cite{radford2021CLIP}) tend to be similar, making it difficult for the diffusion model to distinguish subtle interaction differences.

To capture such fine-grained contextual distinctions, we represent each modality as a set of \emph{tokens} and employ a \emph{cross-attention} mechanism~\cite{vaswani2017attention} that learns localized correspondences between the denoising pose features and conditional modality tokens.
This design enables the model to selectively attend to informative modality tokens, including those implicitly encoding grasp semantics, thereby achieving context-sensitive conditioning.
Specifically, we use modality-specific tokenizers: DistilBERT~\cite{sanh2019distilbert} for the text embedding and the ViT network trained in~\cite{pavlakos:cvpr24:HaMeR} for the image embedding.

\customparagraph{Diffusion process}
We follow the DDPM formulation~\cite{ho:neurips20:ddpm}, where the diffusion process consists of forward and reverse paths spanned with diffusion steps $t\in [0,T]$.
The forward process gradually adds standard Gaussian noise to the input data $\mathbf{x}_0$, which is formulated as
\begin{equation}
    \mathbf{x}_t:=\sqrt{\bar{\alpha}_t} \mathbf{x}_0 + \sqrt{1-\bar{\alpha}_t}\,\epsilon,
\label{eq:forward}
\end{equation}
where $\epsilon \sim \mathcal{N}(0, I)$ is a noise variable and $\alpha_{1:T} \in (0, 1]^T$ is a 
sequence that controls the degree of noise at each timestep $t$.
In contrast, the reverse process learns to iteratively transform noise $\mathbf{x}_T\sim\mathcal{N}(0,I)$ 
toward the target data distribution $\mathbf{x}_0\sim q(x_0)$, defined as
\begin{equation}
p_{\phi}(\mathbf{x}_{0:T}) := p(\mathbf{x}_{T}) \prod_{t=1}^{T} p_{\phi}(\mathbf{x}_{t-1} \mid \mathbf{x}_{t}),
\label{eq:reverse}
\end{equation}
where $p_{\phi}$ is a model distribution defined by the network $f_{\phi}(\cdot)$.
We train the denoising network to approximate the original data $\mathbf{x}_0$ following~\cite{Tevet:iclr23:MDM,lee:cvpr24:interhandgen,Ohkawa:iccv25:SCGen}.

The training loss is computed between the original data $\mathbf{x}_0$ and the generated data $\hat{\mathbf{x}}_0$. 
We take the L2 loss between $\mathbf{x}_0$ and $\hat{\mathbf{x}}_0$ for the pose space, denoted as $L_\theta$. 
After passing to the MANO layer, we compute the L2 loss on the mesh vertices and 3D joint space, denoted as $L_v$ and $L_j$. The overall loss is formulated as
\begin{equation}
L_{total} = \lambda_{\theta}L_\theta+\lambda_v L_v + \lambda_j L_j.
\label{eq:loss}
\end{equation}
$\lambda_\theta$, $\lambda_v$, and $\lambda_j$ are the weights for each loss term.

To improve the robustness of the model, we randomly dropout the conditional input $\mathbf{c_D}$ and $\mathbf{c}_I$, inspired by~\cite{muller:cvpr:24BUDDI,lee:cvpr24:interhandgen,stathopoulos:cvpr:24scorehmr,ho:22:cfg}
. 
Specifically, we randomly mask the affordance description and the RGB image with a probability of 0.1.
This emulates conditional and unconditional generation within a single model.
This strategy brings several benefits: (i) preventing the model from overfitting to a specific conditioning modality, leading to a strong prior learning over 3D hand poses, and 
(ii) improving flexibility at inference time, allowing seamless switching between conditional and unconditional sampling without retraining.

\begin{figure}[!t]
\centering

\begin{minipage}[c]{0.48\textwidth}%
\scriptsize%
\resizebox{0.9\columnwidth}{!}{%
\begin{tabular}{@{}l@{}}
\begin{minipage}{\columnwidth}
\hrule \vspace{2mm}
\textbf{Algorithm 1: Diffusion-based pose refinement.}
\vspace{1mm} \hrule \vspace{1mm}
\begin{algorithmic}[1]
\Statex{\textit{Diffuse 3D pose and visibility labels $\mathbf{v}$ of $\tilde{\mathbf{x}}_0$}}
\State $\tilde{\mathbf{x}}_{n} \leftarrow \sqrt{\bar{\alpha}_n} \tilde{\mathbf{x}}_0 + \sqrt{1-\bar{\alpha}_n}\,\epsilon$
\State $\mathbf{v} = V(\tilde{\mathbf{x}}_0)$ \hfill {\color{gray}{\textit{// 1: visible, 0: occluded}}}
\For{$t=n$ to 1}
        \State $\hat{\mathbf{x}}_0 \leftarrow f(\tilde{\mathbf{x}}_t, t, \mathbf{c})$ 
        
        \Statex \hspace{0.2cm} \(\triangleright\) \textit{2D keypoint fitting only for visible joints}
        \State $\hat{\mathbf{x}}_{0} \leftarrow \hat{\mathbf{x}}_{0} - \lambda_{2d} \nabla_{\hat{\mathbf{x}}_0} \mathcal{L}_2(M(\hat{\mathbf{x}}_0), M(\tilde{\mathbf{x}}_0)) \odot \mathbf{v}$

        \State $\epsilon_t \leftarrow \frac{1}{\sqrt{1-\bar{\alpha}_t}} (\tilde{\mathbf{x}}_t - \sqrt{\bar{\alpha}_t} \hat{\mathbf{x}}_0)$
        \State $\tilde{\mathbf{x}}_{t-1} \leftarrow \sqrt{\bar{\alpha}_{t-1}} \hat{\mathbf{x}}_0 + \sqrt{1-\bar{\alpha}_{t-1}} \epsilon_t$                
\EndFor
\State \Return $\hat{\mathbf{x}}_0$
\end{algorithmic}
\vspace{1mm} \hrule
\end{minipage}
\end{tabular}%
}
\end{minipage}
\hfill%
\begin{minipage}[c]{0.51\textwidth}%
\includegraphics[width=1.1\linewidth]{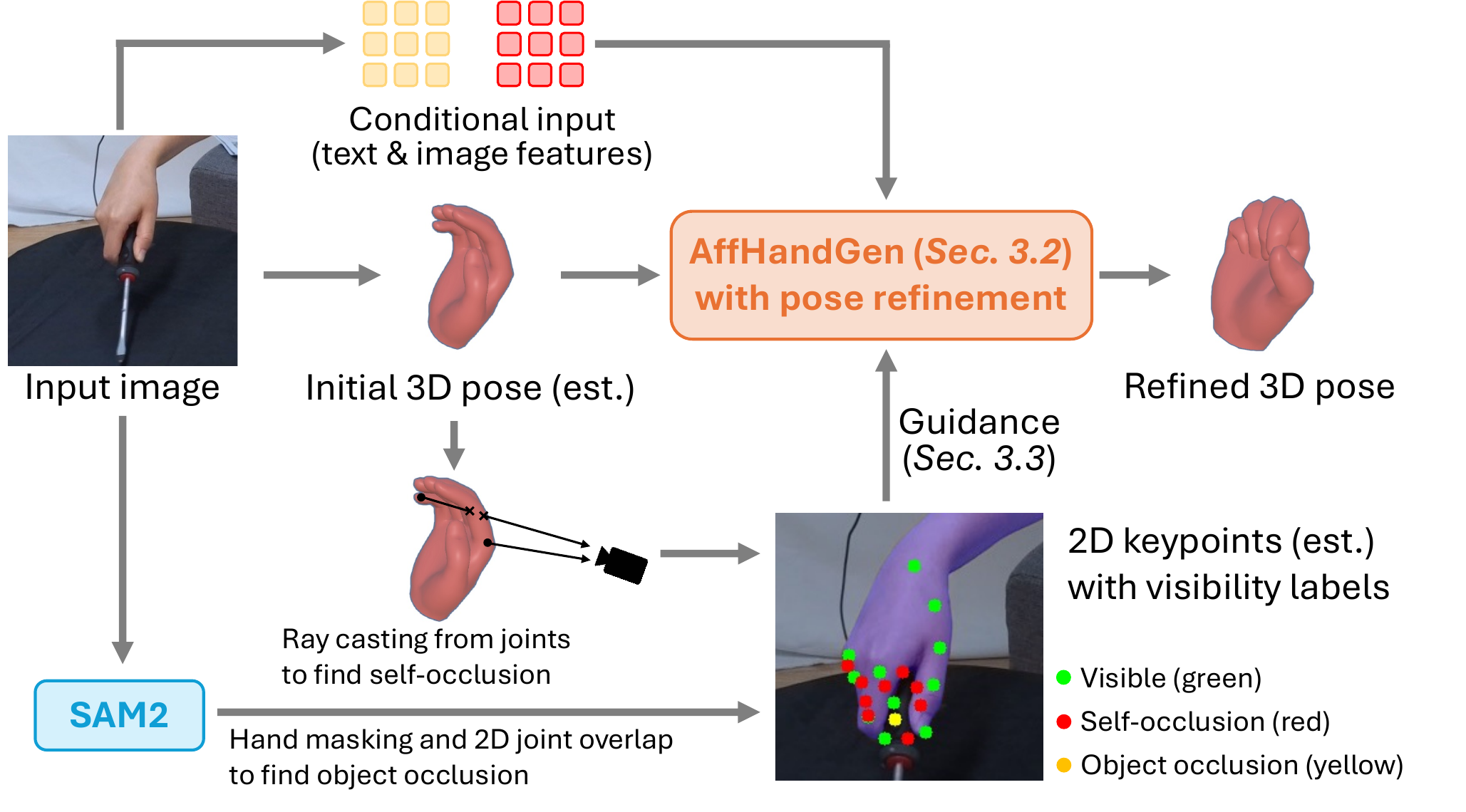}
\end{minipage}

\caption{
\textbf{Diffusion-based pose refinement algorithm.}
Our refinement takes the initial 3D pose estimates $\tilde{\mathbf{x}}_0$ from recent hand reconstruction methods (\eg,~\cite{potamias:cvpr25:wilor}), conditional text and image features $\mathbf{c}$, and detected visibility $V(\cdot)$ in \cref{sec:3d_recon}, and then corrects occluded joints (red and yellow in the right figure) while visible joints (green) remain unchanged.
The visibility labels are obtained from two detection criteria: self-occlusion with ray casting on the MANO mesh and object occlusion with SAM2~\cite{ravi:24:sam2} mask for hands.
This refinement is resumed at the middle of diffusion steps $n$, reducing the total inference cost, and optimized with the 2D keypoint loss based on camera projection function $M(\cdot)$ and its loss weight $\lambda_{2d}$.
}
\label{fig:refinement}
\end{figure}

\subsection{Occlusion-aware prior adaptation to 3D hand reconstruction}\label{sec:3d_recon}
Once the diffusion prior is trained, we adapt the prior to guide and improve 3D hand reconstruction, specifically aiming to refine the 3D pose parameters.
While foundation models for hand reconstruction (\eg, HaMeR~\cite{pavlakos:cvpr24:HaMeR}) effectively align hand meshes with the visible 2D evidence in RGB images, they still struggle to recover accurate 3D poses under severe occlusion, where visual cues are limited.
SCGen~\cite{Ohkawa:iccv25:SCGen} addresses this by applying a 3D human diffusion prior to refine 3D pose estimates with 2D keypoint constraints. 
Given an initial 3D pose $\tilde{\mathbf{x}}_0$ (the target of refinement) and reliable 2D keypoints $\mathbf{y}_0$ detected by human foundation models like Sapiens~\cite{khirodkar:eccv24:sapiens}, it first sets a diffused pose $\tilde{\mathbf{x}}_n$ at the intermediate step $n$ ($0<n<T$) and iteratively denoises the pose until $t=0$, while minimizing the 2D projection loss between $\hat{\mathbf{x}}_0$ and $\mathbf{y}_0$.

Inspired by the diffusion-based refinement for human pose~\cite{Ohkawa:iccv25:SCGen}, we introduce an \emph{affordance-guided diffusion refinement} tailored to 3D hand reconstruction (Fig.~\ref{fig:refinement}).
Our approach newly incorporates contextual cues from textual descriptions and image features, enabling the model to reconstruct hand poses that are physically plausible and semantically consistent with the intended interaction.
Unlike body poses, hands exhibit high articulation and frequent object contacts, leading to more complex occlusion patterns.
To address this, we detect the visibility of each joint and apply selective 2D guidance only to visible joints, allowing the diffusion prior to correct uncertain (occluded) joints through contextual conditioning.

Our refinement framework requires the initial 3D pose $\tilde{\mathbf{x}}_0$, its camera projection (\ie, 2D keypoints) $M(\tilde{\mathbf{x}}_0)$, joint-wise visibility labels $\mathbf{v} \in \{0,1\}^J$ ($J$ is the number of joints), and contextual conditions $\mathbf{c} = \{\mathbf{c}_D,\mathbf{c}_I\}$ from text and image features.
The visibility labels $\mathbf{v}$ are obtained through the detection scheme below and are used to mask the 2D keypoint loss.
Since reliable 2D hand keypoints are more difficult to obtain than those in the body domain~\cite{Ohkawa:iccv25:SCGen}, we instead use the projected 2D keypoints of the initial 3D pose, $M(\tilde{\mathbf{x}}_0)$, along with the visibility labels as supervision signals.
This allows the model to align the pose to confident 2D evidence and denoise uncertain occluded joints based on the contextual diffusion prior.
We further apply the 2D keypoint fitting to the denoised pose $\hat{\mathbf{x}}_0$ as post-processing.
This ensures that the resulting pose after the diffusion-based refinement is well-aligned to the visible 2D keypoints. 

\customparagraph{Occlusion detection}
Estimating joint visibility is useful for downstream HOI tasks, yet challenging.
We employ two criteria to determine the occlusion of hand joints regarding \emph{self-occlusion} and \emph{object occlusion}, resulting in the binary visibility labels $\mathbf{v}$.
For self-occlusion, we determine occluded joints by ray casting given the hand mesh derived from $\tilde{\mathbf{x}}_0$.
A ray is cast from each hand joint toward the camera, and its intersections with the hand surface are computed as~\cite{smith:tog:constraining}.
If the number of intersections is greater than or equal to two, the joint is occluded by the part of hand mesh, regarded as self-occlusion. 
For instance, a ray from the pinky fingertip of \cref{fig:refinement} penetrates into the index finger's mesh, judged as the self-occluded joint.
For object occlusion, we identify object-occluded joints by comparing the projected 2D joints from the initial pose predictions, $M(\tilde{\mathbf{x}}_0)$, and the hand mask obtained from SAM2~\cite{ravi:24:sam2}. 
If the 2D joint position is outside the mask, the joint is regarded as object occlusion.

\section{Experiments}
We present quantitative results on pose refinement, comparing against foundational models for 3D hand reconstruction and diffusion prior baselines with or without using contextual signals. We then present ablations and qualitative analyses of pose refinement and controllable generations. Additional results are found in the supplement.

\customparagraph{Experimental settings} 
We use the HOGraspNet~\cite{cho:eccv24dense} dataset for complex and diverse hand-object interactions, which is the only dataset offering RGB images, 3D assets, and grasp taxonomy labels needed for our task setup.
We follow 28 of the 33 grasp types defined by~\cite{feix:2015:grasptaxonomy} to train a grasp classification network.
We also use the HO3D~\cite{hampali:cvpr20:ho3d} dataset including RGB images and 3D hand-object annotations, as an additional validation and the HInt dataset~\cite{pavlakos:cvpr24:HaMeR} for in-the-wild evaluation.
As these two datasets lack grasp annotations, we apply the grasp classifier trained on \cite{cho:eccv24dense} to assign pseudo grasp labels.
We subsample all datasets and use sparse frames as training and evaluation, 
because consecutive frames sharing similar visuals often contain the same grasp labels and almost identical VLM descriptions.
To ensure diverse contextual signals without heavy semantic overlaps, we resample every 10 frames, resulting in 119K training and 3K evaluation images in \cite{cho:eccv24dense} and 65K training and 1K evaluation images in \cite{hampali:cvpr20:ho3d}.
HOGraspNet's evaluation includes unseen subjects (S1 split~\cite{cho:eccv24dense}) and HO3D comprises unseen views, subjects, and objects.

\begin{table}[t]
\centering
\scriptsize
\setlength{\tabcolsep}{2pt}
\resizebox{\linewidth}{!}{%
\begin{tabular}{lcccccccccc}
\toprule
\multicolumn{11}{c}{\textbf{HOGraspNet}} \\
Method & Avg.$\downarrow$ & Low & Med. & High & PA-MPVPE$\downarrow$ & PCK@5$\uparrow$ & PCK@10$\uparrow$ & PCK@15$\uparrow$ & F@5$\uparrow$ & F@15$\uparrow$ \\
\midrule
HandOccNet~\cite{park:cvpr22:handoccnet} & 15.09 & 14.34 & 15.55 & 17.65 & 12.93 & 12.0 & 42.8 & 67.4 & 0.482 & 0.935 \\
MeshGraphormer~\cite{lin:iccv21:meshgraphormer} & 10.02 & 9.39 & 10.27 & 13.55 & 8.99 & 24.7 & 68.7 & 87.5 & 0.625 & 0.977 \\
HaMeR~\cite{pavlakos:cvpr24:HaMeR} & 8.38 & \underline{7.37} & 8.92 & 12.68 & 7.66 & 38.6 & 78.6 & 91.7 & 0.696 & 0.988 \\
\addlinespace[0.3ex]
\hdashline
\addlinespace[0.3ex]
Hamba~\cite{dong:neurips24:hamba} & 8.46 & 7.44 & 9.00 & 12.72 & 7.57 & 38.3 & 78.9 & 91.8 & 0.698 & 0.987 \\
\quad + \textbf{Ours} & \underline{7.85} & 7.49 & 10.21 & \underline{9.46} & \underline{7.11} & \underline{41.2} & \underline{81.6} & \underline{93.2} & \underline{0.718} & \underline{0.991} \\
\addlinespace[0.3ex]
\hdashline
\addlinespace[0.3ex]
WiLoR~\cite{potamias:cvpr25:wilor} & 8.22 & 7.42 & 8.56 & 12.41 & 7.57 & 39.7 & 79.4 & 92.2 & 0.702 & 0.989 \\
\quad + InterHandGen~\cite{lee:cvpr24:interhandgen} & 8.99 & 7.64 & 9.72 & 14.61 & 7.91 & 34.8 & 75.4 & 90.1 & 0.681 & 0.986 \\
\quad + w/ Full Cond.\,($D$+$I$) & 8.06 & 7.50 & \underline{8.29} & 11.12 & 7.29 & 40.3 & 79.6 & 92.3 & 0.713 & 0.990 \\
\quad + \textbf{Ours} & \textbf{7.53} & \textbf{7.29} & \textbf{7.57} & \textbf{9.37} & \textbf{6.90} & \textbf{44.1} & \textbf{82.5} & \textbf{93.6} & \textbf{0.731} & \textbf{0.992} \\
\midrule
\multicolumn{11}{c}{\textbf{HO3D}} \\
Method & Avg.$\downarrow$ & Low & Med. & High & PA-MPVPE$\downarrow$ & PCK@5$\uparrow$ & PCK@10$\uparrow$ & PCK@15$\uparrow$ & F@5$\uparrow$ & F@15$\uparrow$ \\
\midrule
HandOccNet~\cite{park:cvpr22:handoccnet} & 8.83 & 8.46 & 9.44 & 9.86 & 8.80 & 31.4 & 69.3 & 86.7 & 0.572 & 0.965 \\
MeshGraphormer~\cite{lin:iccv21:meshgraphormer} & 12.55 & 11.71 & 13.15 & 18.79 & 12.83 & 16.6 & 52.0 & 73.3 & 0.436 & 0.908 \\
HaMeR~\cite{pavlakos:cvpr24:HaMeR} & 8.15 & 7.34 & 9.59 & 9.54 & 8.33 & 37.7 & 74.5 & 89.0 & 0.632 & 0.973 \\
\addlinespace[0.3ex]
\hdashline
\addlinespace[0.3ex]
Hamba~\cite{dong:neurips24:hamba} & 7.42 & \textbf{7.10} & 7.88 & 8.71 & 7.66 & \underline{40.3} & 77.9 & 91.3 & \underline{0.655} & \underline{0.982} \\
\quad + \textbf{Ours} & \textbf{7.29} & 7.25 & \textbf{7.14} & 8.39 & \textbf{7.37} & \textbf{41.1} & \textbf{78.9} & \underline{91.9} & \textbf{0.657} & \textbf{0.983} \\
\addlinespace[0.3ex]
\hdashline
\addlinespace[0.3ex]
WiLoR~\cite{potamias:cvpr25:wilor} & 7.43 & 7.19 & 7.85 & \underline{7.87} & 7.63 & 39.0 & 77.9 & \underline{91.9} & 0.650 & \textbf{0.983} \\
\quad + InterHandGen~\cite{lee:cvpr24:interhandgen} & 7.55 & 7.30 & 7.99 & 7.99 & 7.64 & 38.1 & 77.6 & 91.4 & 0.638 & 0.980 \\
\quad + w/ Full Cond.\,($D$+$I$) & 7.65 & 7.35 & 8.15 & 8.51 & 7.72 & 37.6 & 77.1 & 91.0 & 0.633 & 0.979 \\
\quad + \textbf{Ours} & \underline{7.37} & \underline{7.11} & \underline{7.83} & \textbf{7.80} & \underline{7.46} & 39.1 & \underline{78.7} & \textbf{92.0} & 0.645 & \textbf{0.983} \\
\bottomrule
\end{tabular}}
\vspace{1mm}
\caption{\textbf{Results of 3D hand pose estimation on HOGraspNet and HO3D.}
We report PA-MPJPE in millimeters for varying occlusion levels: Low (0–5 occluded joints), Medium (6–10), and High (11–21), together with PA-MPVPE in mm, PCK@5/10/15, and F@5/15\,mm.
The diffusion baselines (InterHandGen, w/ Full Cond.) are applied on top of WiLoR, while our refinement is applied on top of both WiLoR and Hamba.
\textbf{Bold} and \underline{underline} denote the best and second-best.}
\label{tab:main_res_ho_graspnet_ho3d}
\end{table}

For baselines of 3D hand reconstruction, we test a graph-based transformer, \textbf{MeshGraphormer}~\cite{lin:iccv21:meshgraphormer}, an occlusion-robust network, \textbf{HandOccNet}~\cite{park:cvpr22:handoccnet}, two foundational transformers, \textbf{\mbox{HaMeR}}~\cite{pavlakos:cvpr24:HaMeR} and \textbf{\mbox{WiLoR}}~\cite{potamias:cvpr25:wilor}, and a Mamba-based model, \textbf{Hamba}~\cite{dong:neurips24:hamba}.
For the diffusion prior, we train a 3D hand diffusion model, \textbf{\mbox{InterHandGen}}~\cite{lee:cvpr24:interhandgen}, by adapting its architecture from two-hand input to a single-hand input for this task, which lacks any contextual conditions.
For a fair comparison with our method, we further implement \textbf{\mbox{InterHandGen w/ Full Cond.}} modified to take the text and image conditions as global feature vectors.
These diffusion baselines are trained from scratch and applied to refine 3D poses from WiLoR with the refinement method in \cref{sec:3d_recon}.

We evaluate reconstruction performance using both 3D and 2D metrics.
For 3D joints and mesh vertices, we report Procrustes-aligned mean errors (PA-MPJPE and PA-MPVPE).
We also report mesh F-scores under 5\,mm and 15\,mm thresholds (F@5 and F@15), which evaluate surface reconstruction accuracy within the corresponding distance thresholds.
For 2D joint accuracy, we report PCK@5, PCK@10, and PCK@15, which measure the percentage of keypoints whose 2D errors fall within 5, 10, and 15 pixels, respectively.
To measure the model's robustness under increasing ambiguities, we further provide evaluation subsets according to different occlusion levels (Low, Medium, High), grouped by the number of visible joints detected in \cref{sec:3d_recon}.

\customparagraph{Implementation details}
For generation, our attention layers follow a latent size of 512 and four heads.
The loss weights are set to $\lambda_\theta=1$, $\lambda_v=10000$, and $\lambda_j=10000$.
We set the number of tokens for text and image embeddings to 64 and 192, respectively. 
We train diffusion models for 80 epochs with a batch size of 64 using an Adam optimizer~\cite{kingma:iclr15:adam} with a learning rate of 1e-3.
We set the maximum diffusion steps to $T=1000$ and use a cosine scheduler~\cite{nichol:icml21:improved} for the forward process. For refinement, we implement faster sampling techniques.
We adaptively set the start refinement step $n \in [n_{min}, n_{max}]$ to the value chosen linearly according to the number of occluded joints.
Larger occlusion counts result in larger values of $n$ that require more refinement steps, conversely setting fewer steps in less occluded samples.
We also sparsely sample every $m$ step in the reverse process. 
On HOGraspNet, we set $\lambda_{2d}=0.005$, $n=[100, 1000]$, $m=20$. 
HO3D requires fewer refinement steps and increased 2D fitting as the initial predictions are more reliable: we set $\lambda_{2d}=0.5$, $n=[10, 20]$, $m=3$. 
Additional analysis of hyperparameter sensitivity is found in the supplement.

\subsection{Results}
\customparagraph{3D hand pose estimation}
\cref{tab:main_res_ho_graspnet_ho3d} reports the 3D pose evaluation on HOGraspNet~\cite{cho:eccv24dense} and HO3D~\cite{hampali:cvpr20:ho3d}. Recent foundation models outperform the CNN-based HandOccNet~\cite{park:cvpr22:handoccnet}, yet the high-occlusion regime remains challenging ($\sim$12\,mm error on the ``High'' subset).
The diffusion baseline InterHandGen~\cite{lee:cvpr24:interhandgen} barely corrects poses without contextual signals, and even ``w/ Full Cond.'' (image and text as global vectors) only partially improves the ``Med.''/``High'' splits over WiLoR~\cite{potamias:cvpr25:wilor}.
In contrast, our prior consistently improves WiLoR across \emph{all} occlusion levels, reaching an average error of 7.53\,mm, with a substantial 3.04\,mm (24\%) reduction on the ``High'' occlusion subset. This gain comes from fine-grained cross-attention to image and text tokens and occlusion-aware refinement, rather than attending conditions as coarse global vectors like ``IHGen w/ Full Cond.''
The same trend holds on HO3D: although WiLoR and Hamba are already highly competitive, our prior refines both without degradation on average, achieving 7.37\,mm with WiLoR and the best 7.29\,mm with Hamba. This confirms that the gains are not specific to a single base estimator, but generalize across different estimators.
\begin{table}[t]
\centering
\scriptsize
\setlength{\tabcolsep}{3pt}
\begin{minipage}[t]{0.40\linewidth}
\centering
\resizebox{\linewidth}{!}{%
\begin{tabular}{lc|c!{\vdashrule}ccc}
  \toprule
  \textbf{(a) Text type} & \textbf{RGB?} & \textbf{Avg.} & \textbf{Low} & \textbf{Med.} & \textbf{High}  \\
  \midrule
  (1) N/A & - & 9.28 & 8.16 & 9.82 & 14.6 \\
  (2) N/A & \checkmark & 7.90 & 7.69 & 7.89 & 9.91 \\
  (3) Simple VLM & - & 8.67 & 8.00 & 9.03 & 11.4 \\
  (4) Simple VLM & \checkmark & 7.76 & 7.64 & 7.70 & 9.50 \\
  (5) Aff. & - & 8.43 & 7.62 & 8.90 & 11.5  \\
  \rowcolor{black!25}
  (6) Aff. & \checkmark & \underline{7.53} & \underline{7.29} & \underline{7.57} & \underline{9.37}  \\
  \addlinespace[1pt]
  \cdashline{1-6}
  \addlinespace[1pt]
  (7) Aff. w/ GT grasp  & \checkmark & \textbf{7.51} & \textbf{7.28} & \textbf{7.54} & \textbf{9.35}   \\
  \bottomrule
\end{tabular}}
\end{minipage}
\hfill
\begin{minipage}[t]{0.21\linewidth}
\centering
\resizebox{\linewidth}{!}{%
\begin{tabular}{lcc}
\toprule
\textbf{(b) Text input} & Avg.$\downarrow$ & FPS$\uparrow$ \\
\midrule
\rowcolor{black!25} (1) Qwen + Mistral & \textbf{7.53} & 0.31 \\
(2) Qwen keywords   & 7.59          & 0.39 \\
(3) Grasp labels    & 7.69          & 1242 \\
(4) Qwen (one-shot) & 7.68          & 1.19 \\
(5) Aff.\ classifier $^{*}$ & 7.57  & 96   \\
\midrule
\textbf{Seg.\ mask} & Avg.$\downarrow$ & FPS$\uparrow$ \\
\midrule
\rowcolor{black!25} (6) SAM2~\cite{ravi:24:sam2} & \textbf{7.53} & 8.1  \\
(7) InterFormer~\cite{su:iclr26:interformer}    & 7.54          & 16.3 \\
(8) None            & 7.55          & ---  \\
\bottomrule
\end{tabular}}
\end{minipage}
\hfill
\begin{minipage}[t]{0.17\linewidth}
\centering
\resizebox{\linewidth}{!}{%
\begin{tabular}{lc}
\toprule
\textbf{(c) 2D guid.} & Avg.$\downarrow$ \\
\midrule
\rowcolor{black!25} $\lambda_{2d}{=}0.005$ & \textbf{7.53} \\
$\lambda_{2d}{=}0$     & 7.65 \\
\midrule
\textbf{V+L enc.} & Avg.$\downarrow$ \\
\midrule
\rowcolor{black!25} HaMeR+DistilBERT & \textbf{7.53} \\
CLIP             & 11.36 \\
\bottomrule
\end{tabular}}
\end{minipage}
\hfill
\begin{minipage}[t]{0.17\linewidth}
\centering
\resizebox{\linewidth}{!}{%
\begin{tabular}{lc}
\toprule
\textbf{(d) Aff.\ field} & Avg.$\downarrow$ \\
\midrule
\rowcolor{black!25} Full   & \textbf{7.53} \\
w/o obj.\ category & 7.53 \\
w/o interaction    & 7.54 \\
w/o intention      & 7.54 \\
w/o obj.\ shape    & 7.56 \\
w/o obj.\ size     & 7.59 \\
w/o grasp tax.\ & 7.71 \\
\bottomrule
\end{tabular}}
\end{minipage}

\vspace{3mm}
\caption{\textbf{Ablation study on HOGraspNet} (PA-MPJPE).
\textbf{(a)} Conditional modality compares text inputs and RGB image features.
\textbf{(b)}~Text input and visibility masks, \textbf{(c)}~2D keypoint guidance and V+L encoder, and \textbf{(d)}~affordance fields; $^{*}$ denotes a lightweight classifier trained on GT-based affordance elements.
{\setlength{\fboxsep}{1.5pt}\colorbox{black!25}{Shaded rows}} mark our setting.
\textbf{Bold} and \underline{underline} denote the best and second-best.}

\label{tab:cond_modality}
\end{table}

\begin{figure*}[t]
    \centering
    \hfill
    \includegraphics[width=\linewidth]{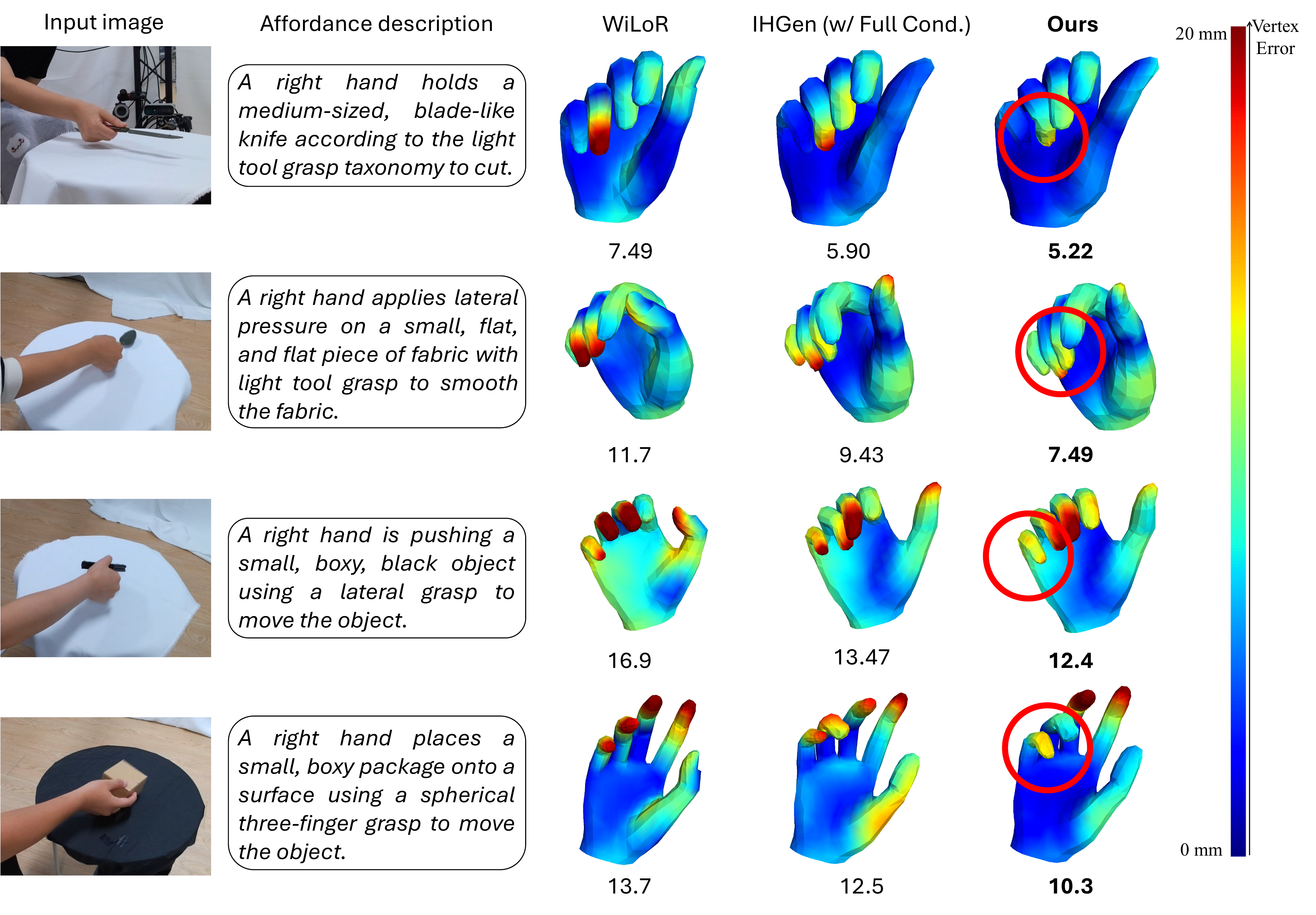}
    \caption{
    \textbf{Qualitative results of diffusion-based refinement on the HOGraspNet dataset~\cite{cho:eccv24dense}.}
    We show refinement results with our diffusion prior and a comparison diffusion baseline, InterHandGen (IHGen)~\cite{lee:cvpr24:interhandgen} with Full Cond.
    The vertex error compared to the ground-truth mesh is color-coded, which is clipped to $[0, 20]\,$mm.
    While the initial hand pose from WiLoR~\cite{potamias:cvpr25:wilor} struggles to estimate joints in occluded regions, our method reasonably refines the estimates to plausibly grasp the object even under occlusion. The values under the samples denote PA-MPJPE.
    }
    \label{fig:refine_qualitative}
\end{figure*}

\customparagraph{Ablation studies}
All ablations are conducted on HOGraspNet and reported in \cref{tab:cond_modality}.
We ablate the main components of our pipeline, including conditional modalities, affordance text generation, visibility masking, 2D keypoint guidance, encoder design, and individual affordance fields.

\customparagraph{(a) Conditional modality} We evaluate the impact of image and text conditioning, comparing the guidance effect from affordance descriptions with simple VLM captions.
The Simple~VLM text is generated by prompting Qwen2.5-VL~\cite{bai2025qwen2.5vl} to describe the given HOI scene (\eg, ``The right hand holds a screwdriver'').
Compared with results of (1,\,3,\,5), the simple caption method does not offer significant improvement over the baseline without text conditioning (N/A), while the affordance descriptions (Aff.) help reduce the ambiguities.
Conditioning image features with ViT encoding (2,\,4,\,6) further proves to be effective, serving to align the predicted pose with visual evidence.
We observe that our configuration (6) performs the best 
7.53\,mm
by utilizing the affordance descriptions and the image features jointly.
We also assess the effect of predicted grasp types (see ``Step~2'' of \cref{sec:desc_gen}) against using ground-truth grasp labels at test time.
We find the results (6,\,7) almost on par, indicating that the predicted grasps in (6) include few failures in classification and do not negatively affect the downstream task.

\noindent\textbf{(b) Text input \& segmentation mask:} For the text input, our two-stage Qwen+Mistral description (1), where the VLM extracts affordance keywords that Mistral summarizes into a coherent sentence, outperforms three simpler alternatives: raw VLM keywords without summarization (2), grasp taxonomy labels alone (3), and a single-shot VLM prompt (4). This confirms that the structured summarization provides a stronger conditioning signal than flat labels or unsummarized text. A lightweight affordance classifier trained with GT-based affordance elements is then used to generate the affordance descriptions (5), recovering most of the accuracy (7.57\,mm) while running at 96\,fps. For the visibility mask, replacing SAM2~\cite{ravi:24:sam2} (6) with the faster InterFormer~\cite{su:iclr26:interformer} (7) doubles throughput at a negligible accuracy cost, and removing the mask entirely (8) only mildly degrades performance.

\noindent\textbf{(c) 2D guidance \& V+L encoder:} 
The 2D keypoint guidance ($\lambda_{2d}{=}0.005$) grounds the refinement to visual evidence during inference, and removing this guidance by setting $\lambda_{2d}{=}0$ degrades the accuracy.
We further ablate the V+L encoder, which denotes the vision and language encoding modules used to condition our diffusion prior.
Our modality-specific encoder, implemented with HaMeR for visual features and DistilBERT for textual affordance descriptions, is also essential to performance.
In contrast, replacing these task-specific encoders with a CLIP encoder substantially degrades the error to 11.36\,mm.
This result suggests that CLIP's global embeddings lack the hand-specific spatial granularity required to distinguish subtle hand-object interactions.

\noindent\textbf{(d) Affordance fields:} Ablating each field of the affordance description shows that the grasp taxonomy contributes the most to the refinement. The object category, shape, size, interaction, and intention also provide complementary gains, highlighting the benefit of combining multiple elements for accurate refinement.

\begin{figure*}[t]
    \centering
    \hfill
    \includegraphics[width=\linewidth]{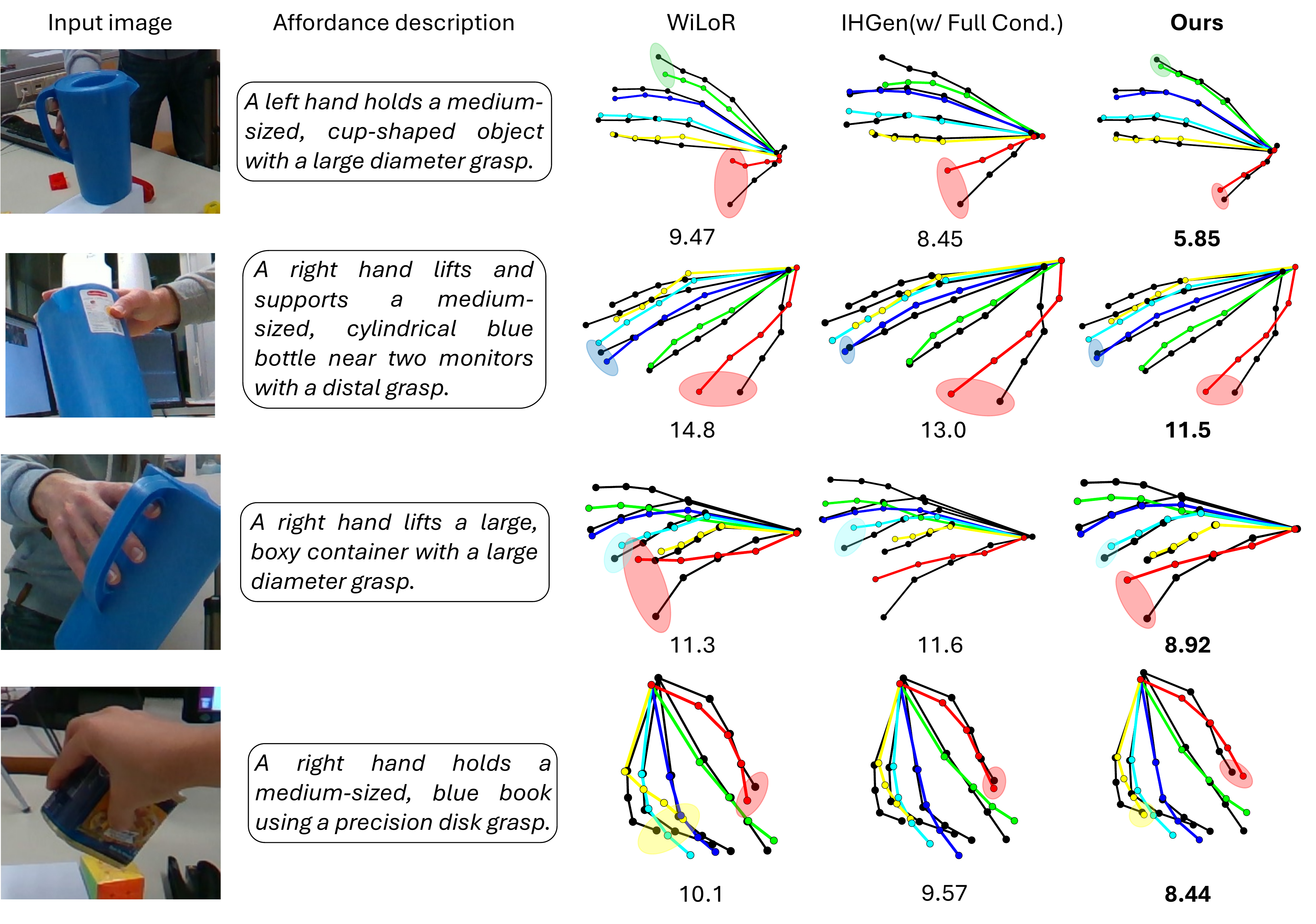}
    \caption{
    \textbf{Qualitative results of diffusion-based refinement on the HO3D dataset~\cite{hampali:cvpr20:ho3d}.}    
    We compare the refinement results of our diffusion prior against a diffusion baseline, InterHandGen (IHGen)~\cite{lee:cvpr24:interhandgen} with Full Cond., both applied to the initial hand poses predicted by WiLoR~\cite{potamias:cvpr25:wilor}. Predicted hand joints are colored by finger, with ground truth displayed in black. Some joint errors are highlighted using colored ellipses. The values under the samples denote PA-MPJPE.
    }
    \label{fig:refine_qualitative_ho3d}
\end{figure*}

\customparagraph{Qualitative results}
Fig.~\ref{fig:refine_qualitative} presents qualitative comparisons on HOGraspNet between the WiLoR predictions, the refined poses by InterHandGen (IHGen) with Full Cond., and our diffusion prior. The initial poses exhibit large errors (red vertices) in the occluded parts (\eg, the ring and pinky fingers are often hidden in the examples).
Our method corrects these errors and improves over the IHGen (w/ Full Cond.) baseline.

Similarly, Fig.~\ref{fig:refine_qualitative_ho3d} visualizes qualitative comparisons on the HO3D dataset. As highlighted by the ellipses, the initial WiLoR predictions suffer from high errors at severely occluded joints. While the diffusion baseline, IHGen (w/ Full Cond.), attempts to correct them, it struggles to effectively reflect the textual affordance descriptions into the refined poses. In contrast, our diffusion prior successfully leverages these text cues, effectively correcting the hidden articulations to form plausible grasps.

\begin{figure}[t]
  \centering
  \begin{minipage}{0.48\textwidth}
    \centering
    \includegraphics[width=\linewidth]{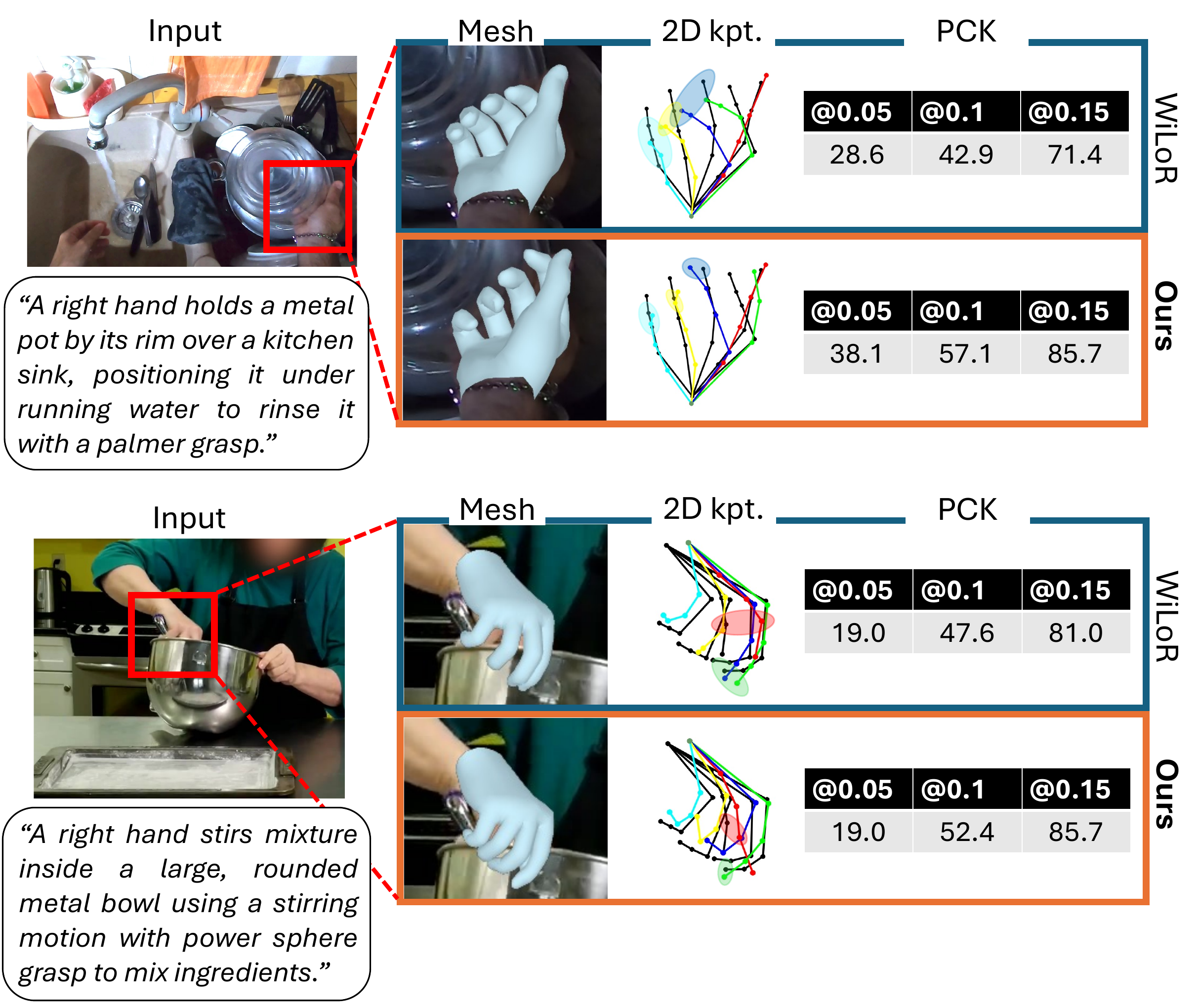}
    \caption{
    \textbf{In-the-wild sample evaluation on the HInt dataset~\cite{pavlakos:cvpr24:HaMeR}.}
    We present PCK (\%) to the 2D keypoint GT (black skeleton).
    The highlighted circle and color in 2D kpt indicate the 2D joint error and its finger type (red circle means thumb error).
    }
    \label{fig:qual_itw}
  \end{minipage}
  \hfill%
  \begin{minipage}{0.5\textwidth}
    \centering
    \includegraphics[width=\linewidth]{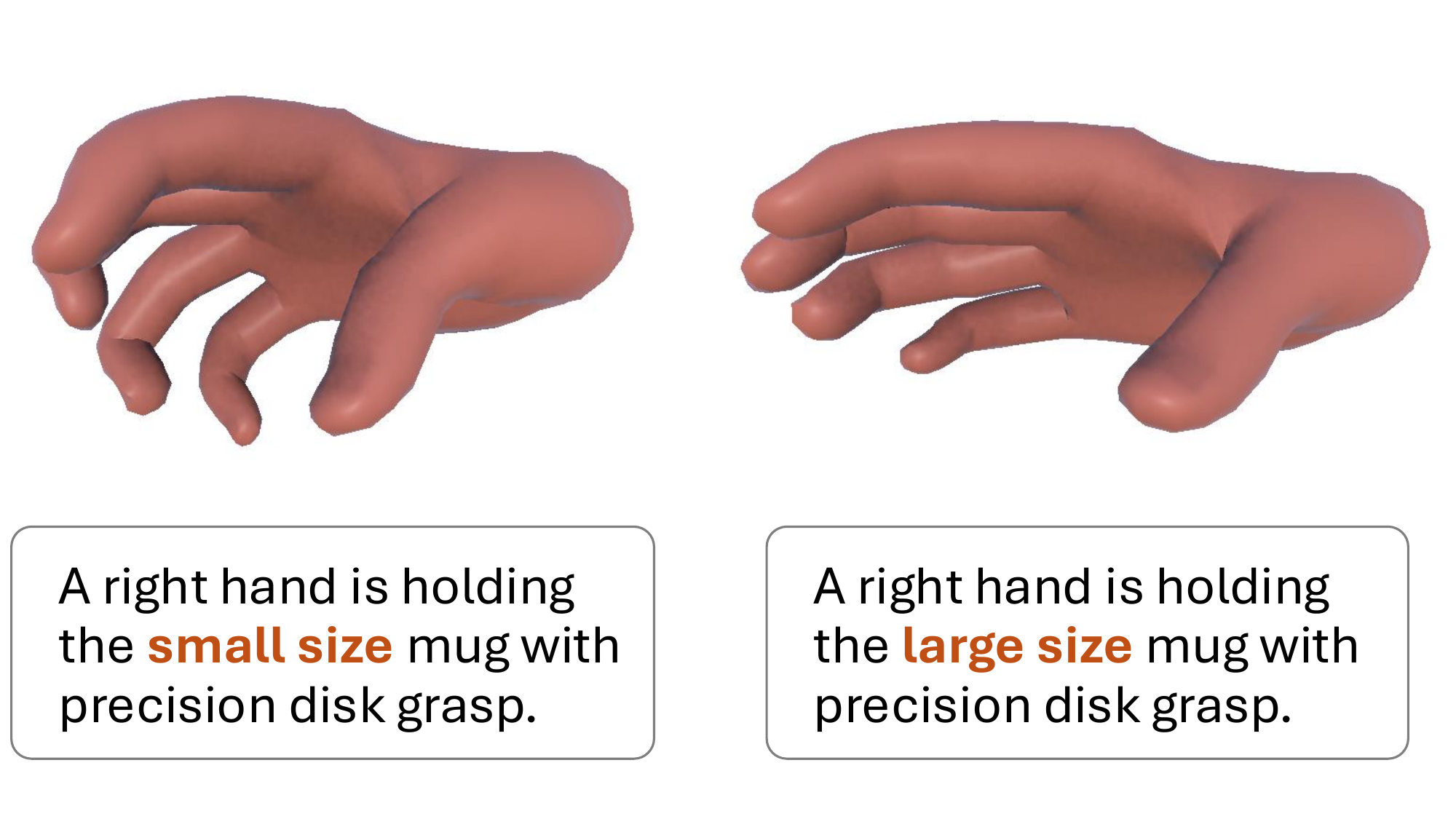}
    \caption{
    \textbf{Qualitative results of our diffusion prior.} 
    We show samples generated from partially different descriptions.
    We find that the diffusion-based generation responds to the change of object size, while remaining consistent with the given description.}
    \label{fig:qual_control}
  \end{minipage}
\end{figure}

Fig.~\ref{fig:qual_itw} shows in-the-wild evaluation on the HInt dataset~\cite{pavlakos:cvpr24:HaMeR}, including 
proxy 2D keypoint evaluation (PCK).
We find stable captioning with generalized VLM capability and successful diffusion-based refinement over WiLoR with improved PCK metrics.
This suggests the applicability of our method in such unconstrained cooking environments where the affordance signals are abundant.

\begin{figure}[t]
  \centering
  \begin{minipage}{0.46\textwidth}
    \centering
        \resizebox{1\linewidth}{!}{%
        \begin{tikzpicture}
        \begin{axis}[
            width=7cm, height=4.5cm,
            xlabel={\footnotesize\textbf{Object category}},
            xlabel shift=-1mm,
            ylabel={\footnotesize\textbf{Size}},
            xtick={0,1,2,3,4},
            xticklabels={cyl., sph., disk, cub., art.},
            x tick label style={rotate=0, anchor=north, font=\scriptsize\bfseries},
            ytick={0,1,2},
            yticklabels={large, med., small}, 
            y tick label style={font=\scriptsize\bfseries},
            enlargelimits=false,
            colorbar,
            colorbar style={
                ylabel={\footnotesize\textbf{Gain (\%)}},
                ylabel style={
                    rotate=180,
                    anchor=north,
                    at={(axis description cs:1,0.5)},
                    yshift=22pt, 
                },
                width=0.2cm,                
                y tick label style={font=\tiny\bfseries,rotate=0},
            },
            colormap name=softgreen,
            point meta min=0,
            point meta max=10,
            every node near coord/.style={
                anchor=center,
                align=center,
                text depth=0pt,
                text=black 
            },
            visualization depends on={value \thisrow{label} \as \mycontent},
            nodes near coords={\mycontent},
        ]
        
        \addplot[
            matrix plot,
            mesh/cols=5,
            mesh/rows=3,
            point meta=\thisrow{improve},
            draw=white,
            line width=0.4pt
        ] table [row sep=\\, col sep=space] {
        x y improve label \\
        0 2 6.97 {{\fontsize{7}{8}\selectfont\textbf{7.43/6.92}} \\ {\fontsize{6}{7}\selectfont\textbf{(-6.97\%)}}} \\
        1 2 3.41 {{\fontsize{7}{8}\selectfont\textbf{7.55/7.29}} \\ {\fontsize{6}{7}\selectfont\textbf{(-3.41\%)}}} \\
        2 2 5.18 {{\fontsize{7}{8}\selectfont\textbf{8.19/7.76}} \\ {\fontsize{6}{7}\selectfont\textbf{(-5.18\%)}}} \\
        3 2 4.19 {{\fontsize{7}{8}\selectfont\textbf{7.19/6.89}} \\ {\fontsize{6}{7}\selectfont\textbf{(-4.19\%)}}} \\
        4 2 6.85 {{\fontsize{7}{8}\selectfont\textbf{8.44/7.87}} \\ {\fontsize{6}{7}\selectfont\textbf{(-6.85\%)}}} \\
        0 1 9.63 {{\fontsize{7}{8}\selectfont\textbf{8.43/7.61}} \\ {\fontsize{6}{7}\selectfont\textbf{(-9.63\%)}}} \\
        1 1 3.55 {{\fontsize{7}{8}\selectfont\textbf{8.26/7.97}} \\ {\fontsize{6}{7}\selectfont\textbf{(-3.55\%)}}} \\
        2 1 6.92 {{\fontsize{7}{8}\selectfont\textbf{8.35/7.77}} \\ {\fontsize{6}{7}\selectfont\textbf{(-6.92\%)}}} \\
        3 1 4.52 {{\fontsize{7}{8}\selectfont\textbf{8.04/7.67}} \\ {\fontsize{6}{7}\selectfont\textbf{(-4.52\%)}}} \\
        4 1 3.95 {{\fontsize{7}{8}\selectfont\textbf{8.56/8.22}} \\ {\fontsize{6}{7}\selectfont\textbf{(-3.95\%)}}} \\
        0 0 3.41 {{\fontsize{7}{8}\selectfont\textbf{8.72/8.42}} \\ {\fontsize{6}{7}\selectfont\textbf{(-3.41\%)}}} \\
        1 0 0.00 {{\fontsize{7}{8}\selectfont\textbf{N/A}}} \\
        2 0 8.35 {{\fontsize{7}{8}\selectfont\textbf{9.43/8.64}} \\ {\fontsize{6}{7}\selectfont\textbf{(-8.35\%)}}} \\
        3 0 3.41 {{\fontsize{7}{8}\selectfont\textbf{7.86/7.60}} \\ {\fontsize{6}{7}\selectfont\textbf{(-3.41\%)}}} \\
        4 0 8.84 {{\fontsize{7}{8}\selectfont\textbf{9.40/8.57}} \\ {\fontsize{6}{7}\selectfont\textbf{(-8.84\%)}}} \\
        };
        
        \end{axis}
        \end{tikzpicture}
        }
        \vspace{-6mm}
        \caption{
        \textbf{Object property-wise performance evaluation on HOGraspNet.}
        The cell value ``WiLoR\,/\,Ours\,(gain\%)'' shows the WiLoR baseline, our method, and the relative gain in PA-MPJPE.
        The x-axis represents cylinder, sphere, disk, cuboid, and articulated shapes.
        }
        \label{fig:obj_eval}
  \end{minipage}
  \hfill
  \begin{minipage}{0.5\textwidth}
    \centering    
        \resizebox{1\linewidth}{!}{%
        \small
        \setlength{\tabcolsep}{8pt}
        \begin{tabular}{lccc}
        \toprule
        Method & Unit & GFLOPs$\downarrow$ & \begin{tabular}[c]{@{}c@{}}Throughput$\uparrow$ \\ (frame/sec)\end{tabular} \\
        \midrule
        WiLoR 
        & per sample 
        & 138.06  
        & 21.0 \\
        
        \cdashline{1-4}[0.8pt/2pt]
        \addlinespace[2pt]
        
        AffHandGen 
        & per diff. step 
        & 0.99 
        & 389.1 \\
        
        AffHandGen 
        & per sample ($\approx$8 steps)
        & \textbf{7.94} 
        & \textbf{48.6} \\
        \bottomrule
        \end{tabular}%
        }
        \captionof{table}{
        \textbf{Computational cost evaluation on HOGraspNet.}
        We measure the inference burden of foundational transformers and our diffusion model.
        As it sparsely samples diffusion steps in inference, our method only requires on average approximately 8 diffusion steps per sample, resulting in significantly lower GFLOPs and higher throughput compared to WiLoR.
        }
        \label{tab:efficiency}
\end{minipage}

\end{figure}

\customparagraph{Controllability of diffusion-based generation}
Fig.~\ref{fig:qual_control} presents qualitative examples of our diffusion model given different textual descriptions.
We validate how pose generation responds to different affordance elements with the rest of the description fixed.
Specifically, we alter only the object size (small \vs large), demonstrating that our model can produce hand poses that adapt to the size change, such as wider grasps for large objects and tighter grasps for small ones.
This suggests that our diffusion prior can faithfully incorporate actionable semantics embedded in the affordance descriptions.

\customparagraph{Object-property-wise evaluation}
We provide further granular evaluations on \cite{cho:eccv24dense} with different object types and sizes. 
As shown in \cref{fig:obj_eval}, while WiLoR suffers large errors on medium cylinders, large disks, and articulated shapes, our method consistently achieves robust refinement across these challenging categories.

\customparagraph{Inference cost}
Owing to our adaptive start steps in refinement ($n$ in \cref{fig:refinement}) and sparse time sampling, our method does not worsen the inference burden in GFLOPs and Throughput compared to the foundation model WiLoR in \cref{tab:efficiency}.

\section{Conclusion}
We present a novel diffusion prior that leverages affordance descriptions generated with VLM's reasoning and grasp classification.
Our model learns the distribution of plausible and functionally coherent 3D hand poses, and then is adapted to refine 3D pose estimates, enabling robust reasoning about occluded joints and more accurate reconstructions in challenging hand-object interactions.
Our experiments demonstrate that the guidance by affordance descriptions substantially improves the refinement quality over both foundational transformer methods and diffusion priors, with strong gains under severe occlusions. 
Our method also provides controllable and interpretable refinements, aligning pose generation with affordance descriptions. 

We note some limitations that open interesting avenues for future work.
Our method currently focuses on static single-hand reconstruction and does not yet explicitly account for 3D object geometry, relying instead on textual descriptions as a proxy to include the object information. 
Extending the approach to explicitly model 3D object interactions is promising for future work. Furthermore, extending our affordance-conditioned prior from static images to continuous temporal domains, such as egocentric video analysis and human motion synthesis, is a highly promising and exciting direction for modeling dynamic hand-object interactions.
\section*{Acknowledgements}
This work was supported by JST K Program Grant Number JPMJKP25V1 and JST ASPIRE Grant Number JPMJAP2303.

\bibliographystyle{splncs04}
\bibliography{./main, ref/hand, ref/base, ref/prior, ref/motion, ref/affordance}

\clearpage
\appendix
\setcounter{page}{1}

\begin{center}
\textbf{\Large{Affordance-Guided Diffusion Prior for 3D Hand Reconstruction}}\\
\vspace{2mm}
\Large{\textmd{--- Supplementary Material ---}}
\end{center}

\section{Grasp Classification}
\begin{figure}[h]
    \centering
    \includegraphics[width=0.65\linewidth]{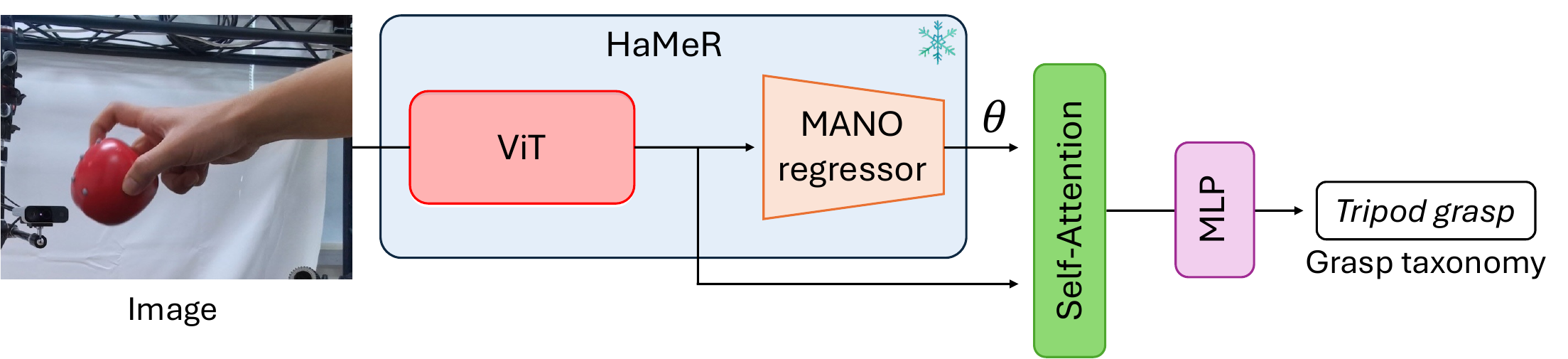}
    \vspace{-2mm}
    \caption{
    Architecture of grasp classification.
    }
    \label{fig:grasp_classification}
\end{figure}

As mentioned in Sec.~\ref{sec:desc_gen}, we adopt a grasp taxonomy classifier to classify grasp types from a single image. The overall architecture for grasp taxonomy classification
is illustrated in Fig.~\ref{fig:grasp_classification}. 
The input image is first encoded by a ViT to obtain visual tokens. 
In parallel, a MANO regressor predicts the hand pose parameters from the encoded visual tokens.
Since grasp taxonomy depends not only on the hand pose but also on the visual context surrounding the hand-object interaction, 
we combine the visual tokens and the MANO pose parameters as a joint input to a self-attention module. We use the ViT encoder and the MANO regressor pretrained on~\cite{pavlakos:cvpr24:HaMeR}.
The output of the self-attention module is then fed into an MLP layer to produce the final grasp taxonomy classification. 
For training, we use the HOGraspNet~\cite{cho:eccv24dense} dataset, adopting the same data size and train/validation split as in the AffHandGen training setup.

\begin{figure}[h]
    \centering
    \includegraphics[width=0.67\linewidth]{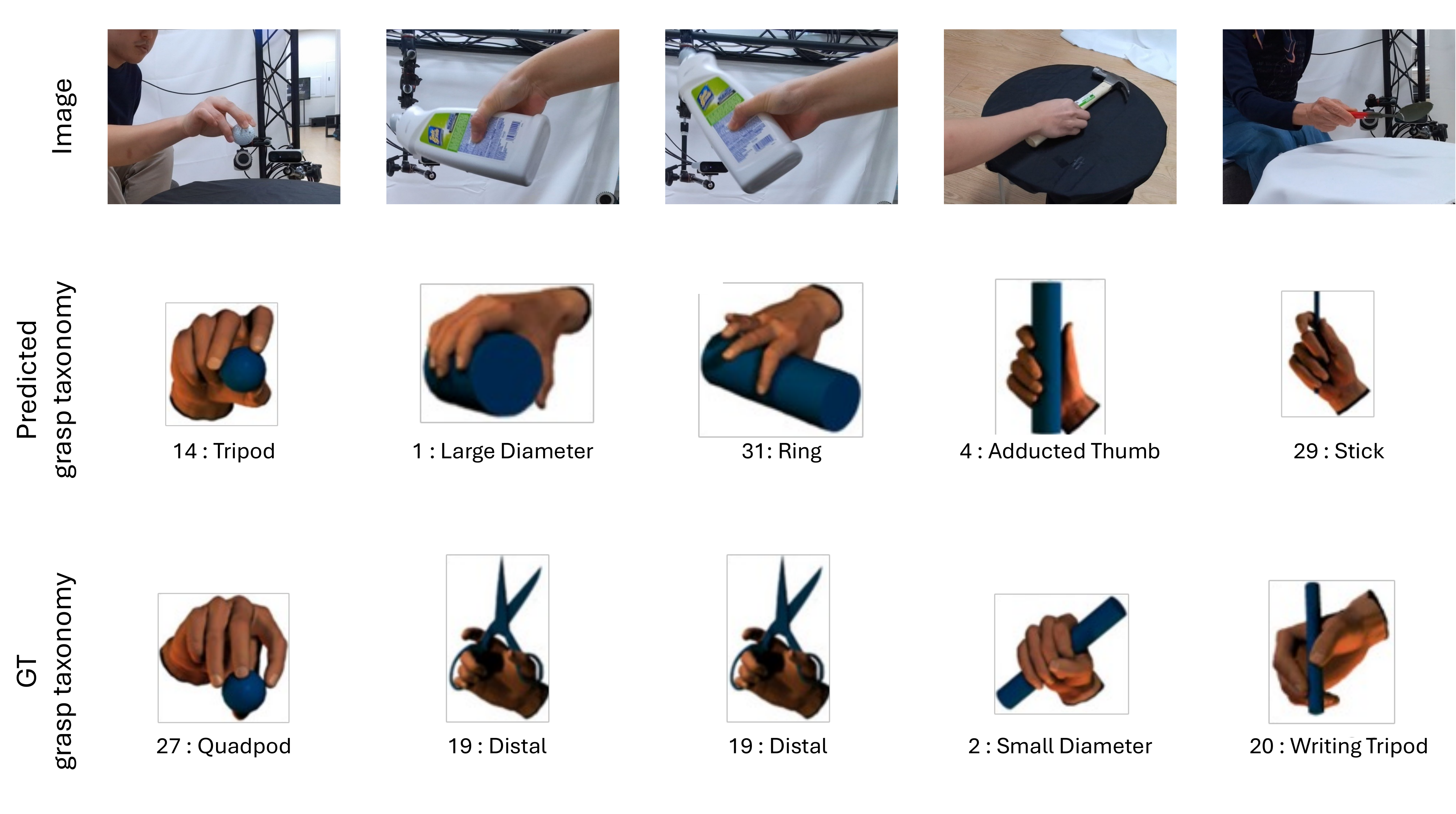}
    \vspace{-4mm}
    \caption{
    Failure cases of grasp classification.
    }
    \label{fig:fail_grasp}
\end{figure}

\begin{figure*}[h]
    \centering
    \begin{minipage}{0.46\textwidth}
    \includegraphics[width=0.95\linewidth]{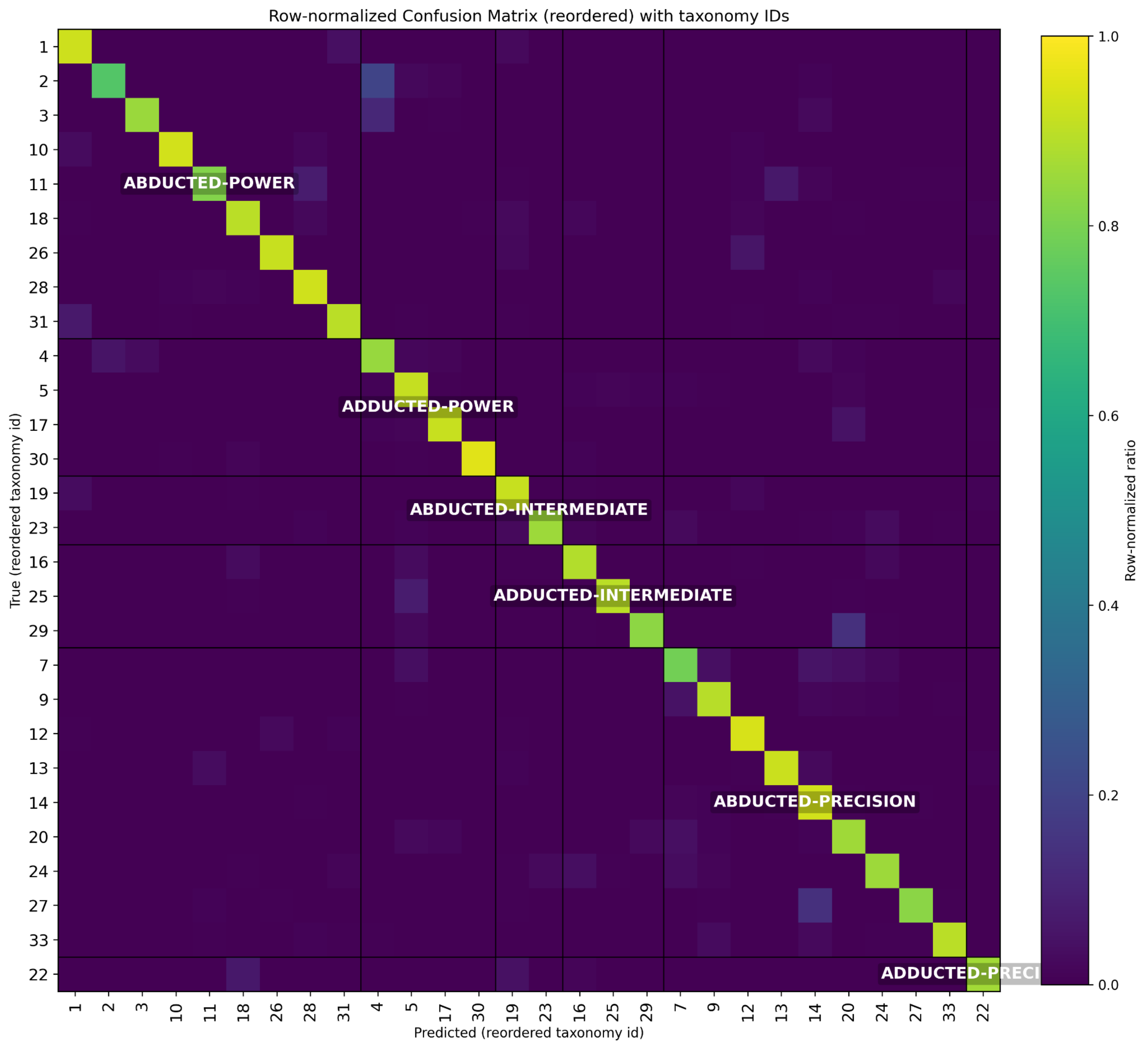}
    \vspace{-2mm}
    \caption{
    Confusion matrix of grasp classification.
    }
    \label{fig:confusion_mat_grasp_classification}
    \end{minipage}
    \hfill
    \begin{minipage}{0.51\textwidth}
    \centering
    \includegraphics[width=0.9\linewidth]{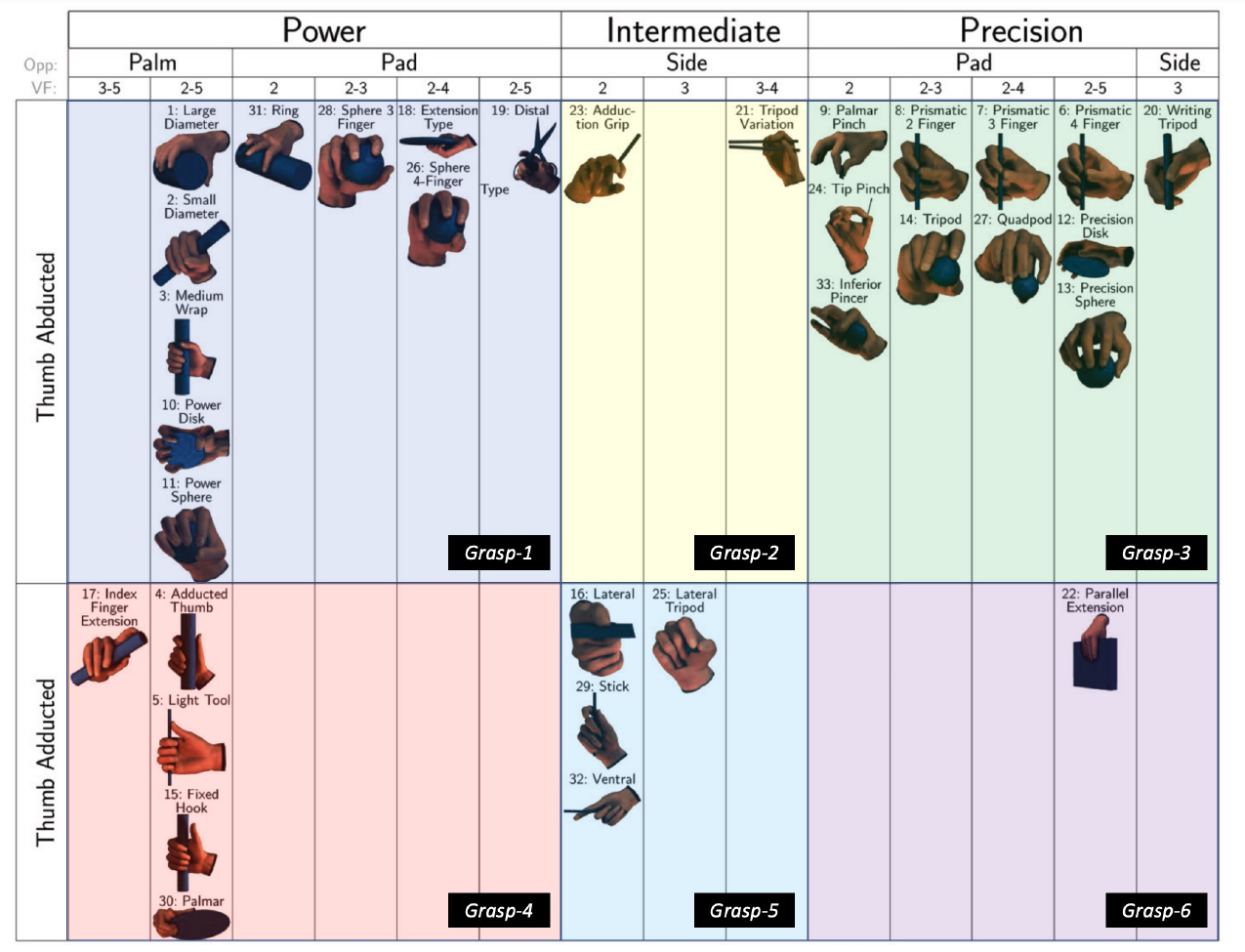}
    \caption{
    The 33 grasp labels~\cite{feix:2015:grasptaxonomy} are grouped into 6 categories. The figure is adapted from~\cite{yu:wacv23:affordanceego}.
    }
    \label{fig:6_grasp}
    \end{minipage}
\end{figure*}

The overall classification accuracy is 87.6\% on the evaluation split.
We show the confusion matrix for grasp taxonomy classification in 
Fig.~\ref{fig:confusion_mat_grasp_classification}, where the original 33 grasp types are grouped into 6 categories based on grasp power and thumb posture, 
following the grouping introduced in~\cite{yu:wacv23:affordanceego} and Fig.~\ref{fig:6_grasp}. 
While predicted taxonomies are mostly accurate, we find some failure examples of our classification model in Fig.~\ref{fig:fail_grasp}. 
These failure cases correspond to grasp types that the model frequently confuses in the confusion matrix. 
Although the predictions are incorrect, the misclassified grasp types are visually and functionally similar. As shown in Tab.~\ref{tab:cond_modality} rows (6) and (7),
the performance gain when replacing predicted grasps with the ground-truth is marginal, indicating that these failures in grasp classification have limited practical impact.

\vspace{-4mm}

\begin{figure}[h]
    \centering
    \includegraphics[width=0.6\linewidth]{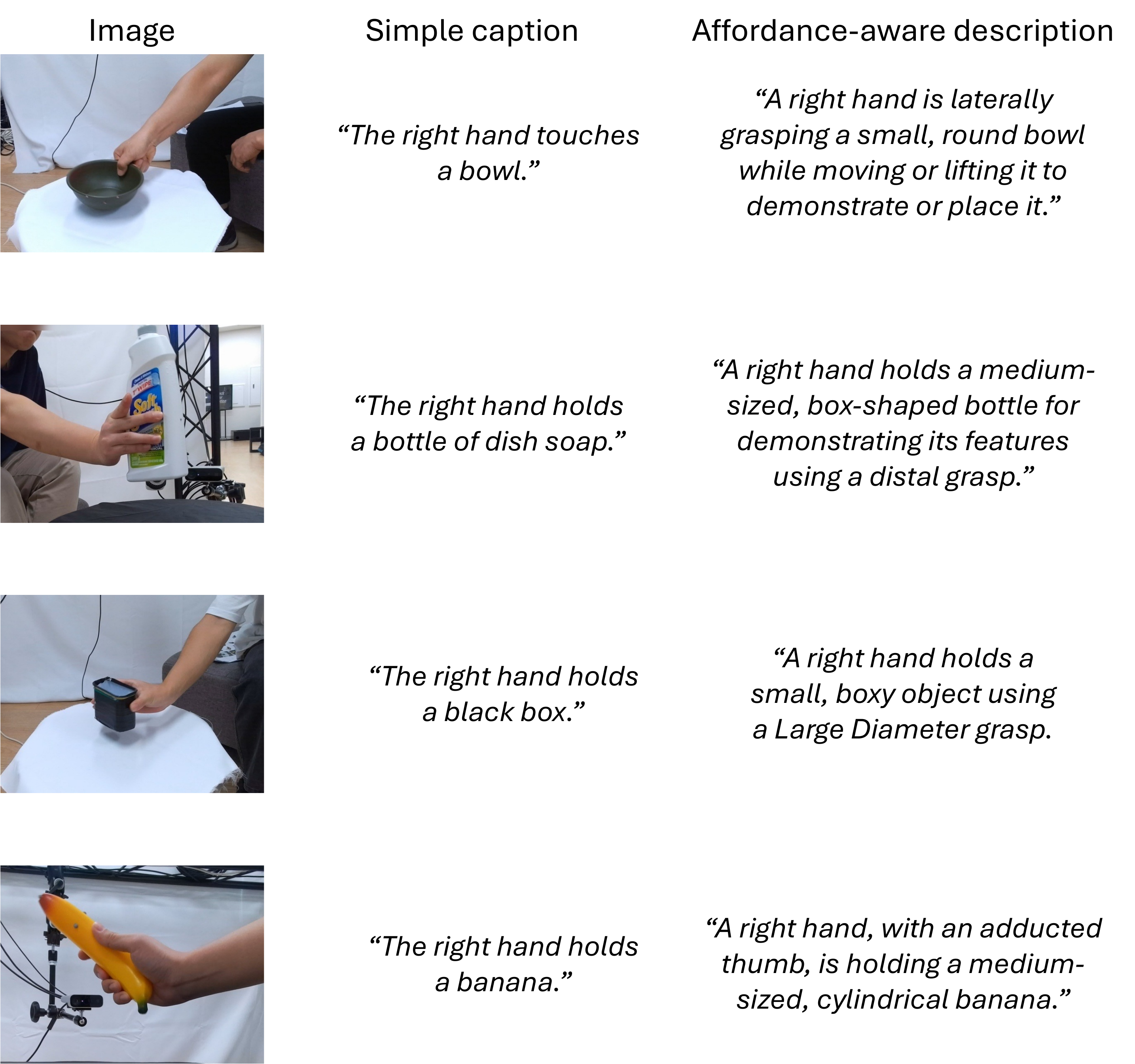}
    \caption{
    Examples of generated simple captions and affordance-aware descriptions.
    }
    \label{fig:caption_generation_examples}
\end{figure}
\vspace{-4mm}

\newpage
\section{Description Generation}
We design two types of descriptions in Sec.~\ref{sec:desc_gen}: a simple caption and an affordance-aware description. These descriptions are generated using a VLM (Qwen2.5-VL~\cite{bai2025qwen2.5vl}).
Examples of both types of generated descriptions are provided in Fig.~\ref{fig:caption_generation_examples}. 
Compared to the simple captions, our affordance-aware descriptions offer more detailed explanations of hand-object interaction, including grasp taxonomy, object size and shape, and other interaction-relevant attributes.

\begin{figure*}[h]
    \centering
    \includegraphics[width=0.76\linewidth]{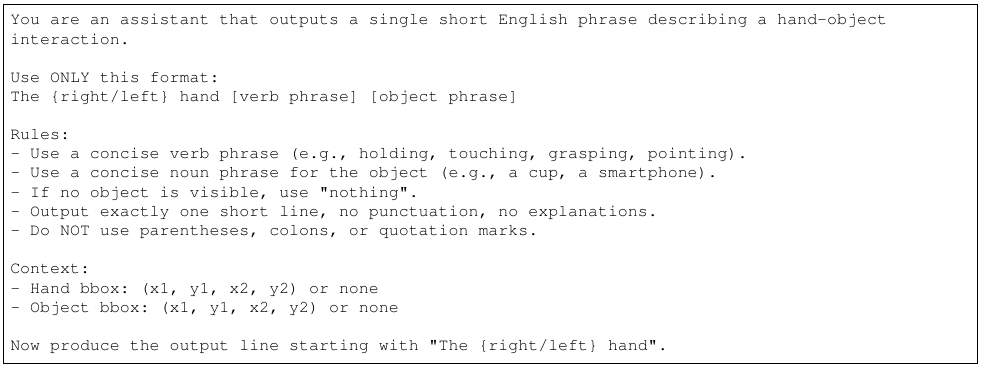}
    \vspace{-2mm}
    \caption{
    Prompt used to generate a simple caption.
    }
    \label{fig:vlm_prompt_simple}
    \vspace{2mm}
    \includegraphics[width=0.76\linewidth]{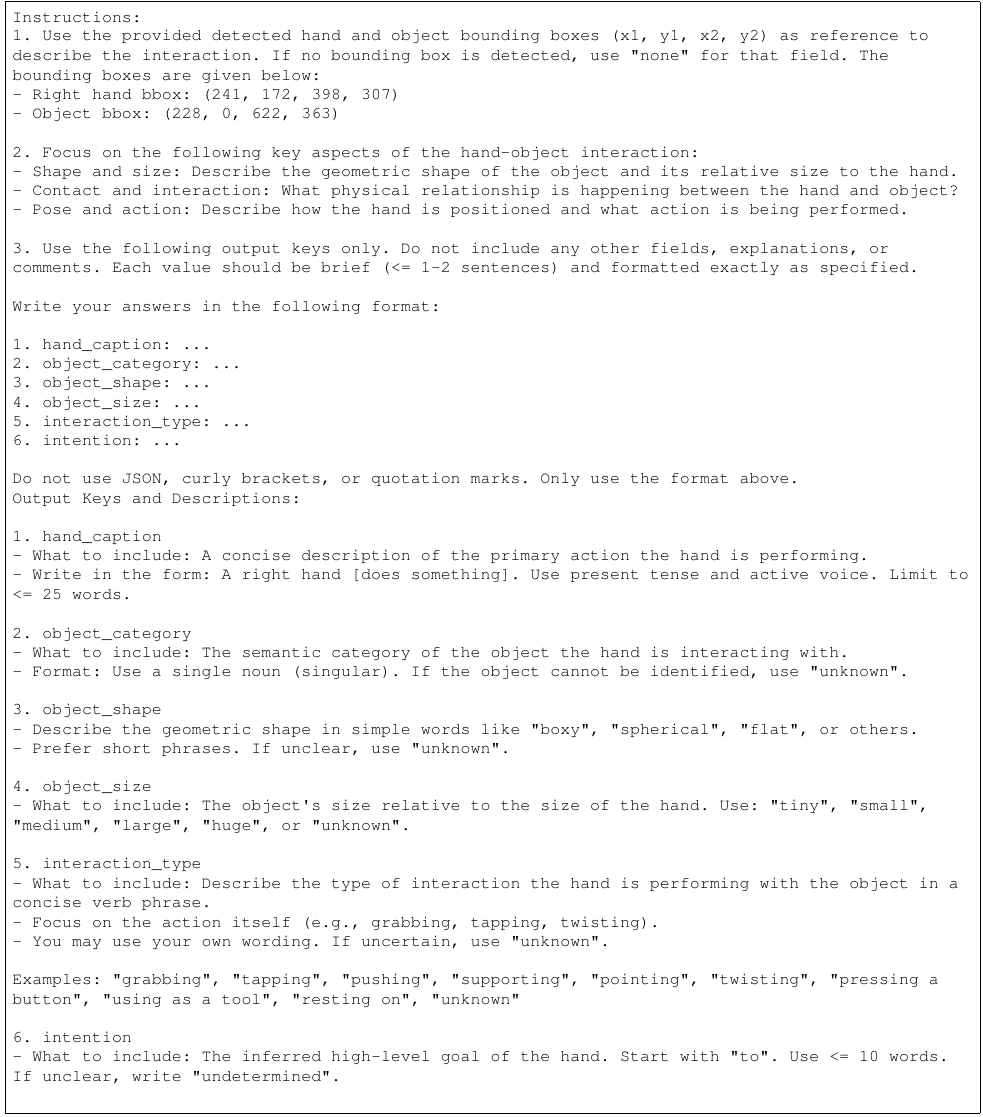}
    \vspace{-2mm}
    \caption{
    Prompt used to generate an affordance-aware description.
    }
    \label{fig:vlm_prompt}
\end{figure*}

For simple caption generation, we feed the image together with the hand-object bounding boxes into the VLM and directly prompt it using the template shown in Fig.~\ref{fig:vlm_prompt_simple}. 
In contrast, the affordance-aware description integrates additional structured cues. 
Specifically, we first feed the image and the hand-object bounding boxes into the VLM with the prompt shown in Fig.~\ref{fig:vlm_prompt}. 
In parallel, we perform grasp classification on the input image and insert the predicted grasp taxonomy into a parsed caption template. 
The combined parsed caption is then provided to an LLM (Mistral-7B~\cite{albert:corr23:mistral7b}), which summarizes it into an affordance-aware description. 
We prompt the LLM with ``Summarize the given parsed captions in a single sentence starting with `A right/left hand\ldots''', and use the resulting summary as our caption.

\newpage

\section{Additional Results}
\begin{figure*}[t]
    \centering
    \includegraphics[width=0.9\linewidth]{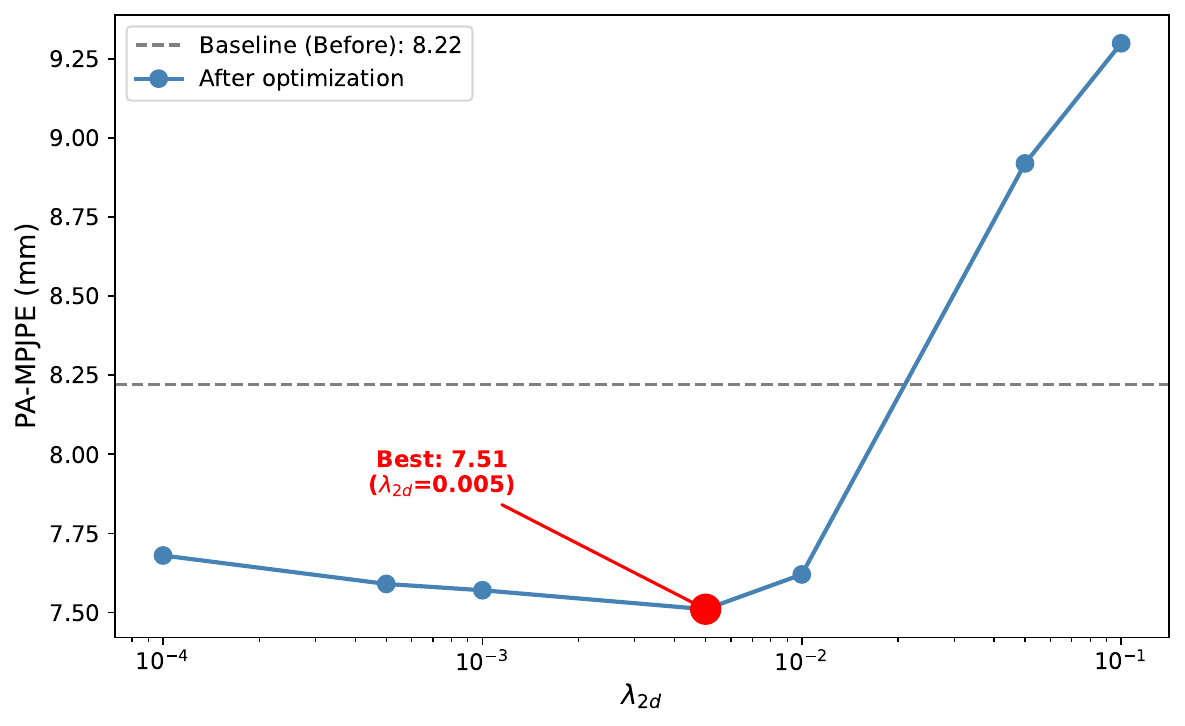}
    \vspace{-2mm}
    \caption{Sensitivity analysis of the 2D reprojection loss weight $\lambda_\text{2d}$ on the HOGraspNet dataset.
    }
    \label{fig:lambda2d_sensitivity}
\end{figure*}

\customparagraph{Hyperparameter sensitivity}
In the refinement stage (Sec.~\ref{sec:3d_recon}), we incorporate a 2D reprojection loss weighted by $\lambda_\text{2d}$ to encourage consistency between the optimized 3D hand pose and 2D keypoint estimates from an off-the-shelf 3D hand reconstruction method. We analyze the model's sensitivity to $\lambda_\text{2d}$ on the HOGraspNet dataset, keeping the maximum diffusion steps $T{=}1000$, start step $n\in[100,1000]$, sampling interval $m{=}20$, and 100 post-processing steps, with image conditioning enabled. As shown in Fig.~\ref{fig:lambda2d_sensitivity}, sweeping $\lambda_\text{2d}$ across four orders of magnitude ($10^{-4}$--$10^{-1}$) reveals a broad plateau of near-optimal performance in the range of $10^{-4}$ to $10^{-2}$. We achieve the best PA-MPJPE of 7.51\,mm at $\lambda_\text{2d}{=}0.005$ (compared to 8.22\,mm before refinement). This broad stability region demonstrates that our refinement pipeline is highly robust and obviates the need for exhaustive hyperparameter tuning in practice.

\begin{figure*}[t]
    \centering
    \includegraphics[width=\linewidth]{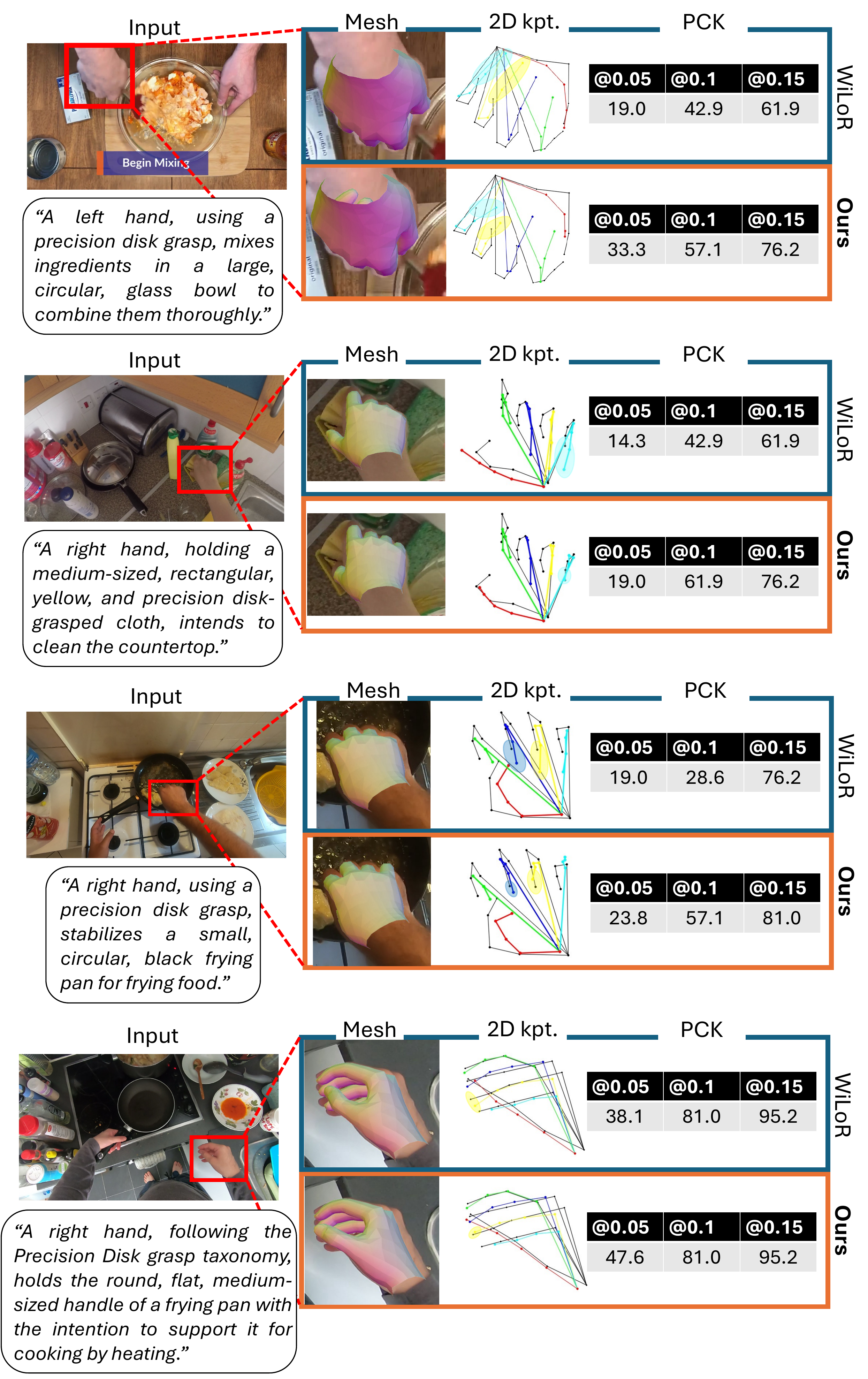}
    \caption{
    Additional qualitative results on in-the-wild images from the HInt dataset~\cite{pavlakos:cvpr24:HaMeR}.
    }
\label{fig:add_res_ITW}
\end{figure*}

\customparagraph{In-the-wild results}
To further assess generalization beyond the training datasets, we evaluate our method on HInt~\cite{pavlakos:cvpr24:HaMeR}, which includes challenging in-the-wild images from the New Days, VISOR, and Ego4D splits. As shown in \cref{tab:hint}, our prior improves WiLoR on occluded joints across all splits and HaMeR on most splits, indicating its effectiveness under unconstrained real-world ambiguities. We further present qualitative results in Fig.~\ref{fig:add_res_ITW}, where our method produces plausible hand poses that remain consistent with the interaction context across diverse real-world scenarios.

\begin{table}[t]
\centering
\scriptsize
\setlength{\tabcolsep}{3pt}
\resizebox{\linewidth}{!}{%
\begin{tabular}{@{}c|lccccccccccc@{}}
\toprule
\multirow{2}{*}{} & \multicolumn{1}{c}{\multirow{2}{*}{Method}} & \multicolumn{3}{c}{New Days} & & \multicolumn{3}{c}{VISOR} & & \multicolumn{3}{c}{Ego4D} \\
& & @0.05$\uparrow$ & @0.1$\uparrow$ & @0.15$\uparrow$ & & @0.05$\uparrow$ & @0.1$\uparrow$ & @0.15$\uparrow$ & & @0.05$\uparrow$ & @0.1$\uparrow$ & @0.15$\uparrow$ \\
\midrule
\parbox[t]{2.5mm}{\multirow{7}{*}{\rotatebox[origin=c]{90}{\color{dark_green}\textbf{All}}}} &
HandOccNet~\cite{park:cvpr22:handoccnet}       &  7.1 & 22.0 & 37.8 & &  5.5 & 18.6 & 33.2 & &  2.9 & 10.2 & 19.6 \\
& MeshGraphormer~\cite{lin:iccv21:meshgraphormer} & 15.0 & 34.7 & 49.1 & & 15.9 & 39.0 & 55.2 & &  7.2 & 19.8 & 31.8 \\
& Hamba~\cite{dong:neurips24:hamba}          & 25.1 & 54.3 & 70.7 & & 19.7 & 49.8 & 69.1 & & 12.0 & 33.0 & 49.0 \\
\cdashline{2-13}
& HaMeR~\cite{pavlakos:cvpr24:HaMeR}          & \textbf{27.7} & 54.3 & 68.9 & & \textbf{25.7} & 55.2 & 71.3 & & \textbf{18.3} & 41.6 & 56.4 \\
& \quad + \textbf{Ours} & 24.6 & 52.6 & 68.5 & & 24.3 & \textbf{55.8} & 72.8 & & 17.5 & \textbf{42.3} & \textbf{58.0} \\
\cdashline{2-13}
& WiLoR~\cite{potamias:cvpr25:wilor}          & 24.9 & \textbf{54.9} & 71.6 & & 21.9 & 53.5 & 72.6 & & 12.5 & 34.2 & 50.3 \\
& \quad + \textbf{Ours} & 23.6 & 54.5 & \textbf{72.8} & & 21.7 & 54.9 & \textbf{74.8} & & 12.5 & 35.6 & 52.5 \\
\midrule
\parbox[t]{2.5mm}{\multirow{7}{*}{\rotatebox[origin=c]{90}{\color{dark_green}\textbf{Visible}}}} &
HandOccNet~\cite{park:cvpr22:handoccnet}       &  8.4 & 25.6 & 43.0 & &  6.2 & 19.8 & 34.7 & &  3.1 & 10.7 & 20.5 \\
& MeshGraphormer~\cite{lin:iccv21:meshgraphormer} & 18.0 & 40.3 & 55.0 & & 21.9 & 49.5 & 64.6 & &  9.4 & 24.1 & 36.1 \\
& Hamba~\cite{dong:neurips24:hamba}          & 30.7 & 63.5 & 78.8 & & 25.7 & 60.9 & 79.0 & & 14.7 & 39.5 & 56.3 \\
\cdashline{2-13}
& HaMeR~\cite{pavlakos:cvpr24:HaMeR}          & \textbf{33.2} & 62.3 & 76.1 & & \textbf{33.0} & \textbf{65.5} & 79.1 & & \textbf{23.6} & 49.3 & 62.6 \\
& \quad + \textbf{Ours} & 29.3 & 60.5 & 75.8 & & 29.9 & 64.6 & 79.8 & & 21.2 & \textbf{49.4} & \textbf{63.7} \\
\cdashline{2-13}
& WiLoR~\cite{potamias:cvpr25:wilor}          & 30.0 & \textbf{63.6} & 79.9 & & 27.4 & 63.9 & 81.8 & & 15.4 & 40.7 & 57.4 \\
& \quad + \textbf{Ours} & 28.0 & 62.6 & \textbf{80.5} & & 26.0 & 63.9 & \textbf{83.0} & & 14.8 & 41.2 & 58.4 \\
\midrule
\parbox[t]{2.5mm}{\multirow{7}{*}{\rotatebox[origin=c]{90}{\color{dark_green}\textbf{Occluded}}}} &
HandOccNet~\cite{park:cvpr22:handoccnet}       &  4.6 & 16.1 & 30.1 & &  4.7 & 17.0 & 31.1 & &  2.6 &  9.3 & 18.0 \\
& MeshGraphormer~\cite{lin:iccv21:meshgraphormer} &  8.4 & 24.1 & 39.1 & &  8.3 & 26.7 & 45.1 & &  4.5 & 14.6 & 26.7 \\
& Hamba~\cite{dong:neurips24:hamba}          & 14.6 & 41.0 & 60.6 & & 12.6 & 36.9 & 58.6 & &  8.6 & 25.3 & 41.1 \\
\cdashline{2-13}
& HaMeR~\cite{pavlakos:cvpr24:HaMeR}          & \textbf{17.2} & \textbf{42.1} & 59.6 & & 16.0 & 43.0 & 62.7 & & 11.6 & 32.5 & 49.4 \\
& \quad + \textbf{Ours} & 15.9 & 40.4 & 59.2 & & \textbf{16.6} & \textbf{45.4} & 65.1 & & \textbf{12.6} & \textbf{33.7} & \textbf{51.5} \\
\cdashline{2-13}
& WiLoR~\cite{potamias:cvpr25:wilor}          & 15.4 & \textbf{42.1} & 61.2 & & 15.4 & 41.3 & 62.9 & &  9.0 & 26.6 & 42.5 \\
& \quad + \textbf{Ours} & 15.1 & \textbf{42.1} & \textbf{63.0} & & 16.3 & 44.3 & \textbf{66.1} & &  9.5 & 28.6 & 45.5 \\
\bottomrule
\end{tabular}}
\caption{\textbf{Quantitative evaluation on the HInt benchmark}~\cite{pavlakos:cvpr24:HaMeR} (PCK@0.05/0.1/0.15).
On occluded joints, our affordance-guided prior improves WiLoR across all three splits (New Days, VISOR, Ego4D) and HaMeR on most of them.}
\vspace{-4mm}
\label{tab:hint}
\end{table}

\begin{figure*}[!t]
    \centering
    \includegraphics[width=0.9\linewidth]{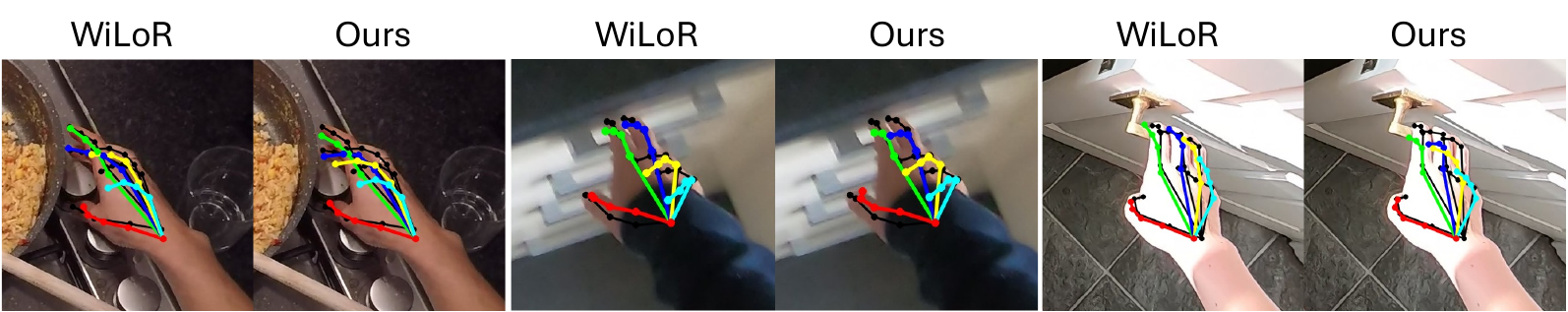}
    \caption{\textbf{Failure cases on HInt.}
    When the interaction deviates from the training distribution, the affordance prior provides weaker refinement over the initial WiLoR estimate.}
    \label{fig:hint_fail}
    \vspace{-3mm}
\end{figure*}

\begin{figure*}[t]
    \centering
    \includegraphics[width=\linewidth]{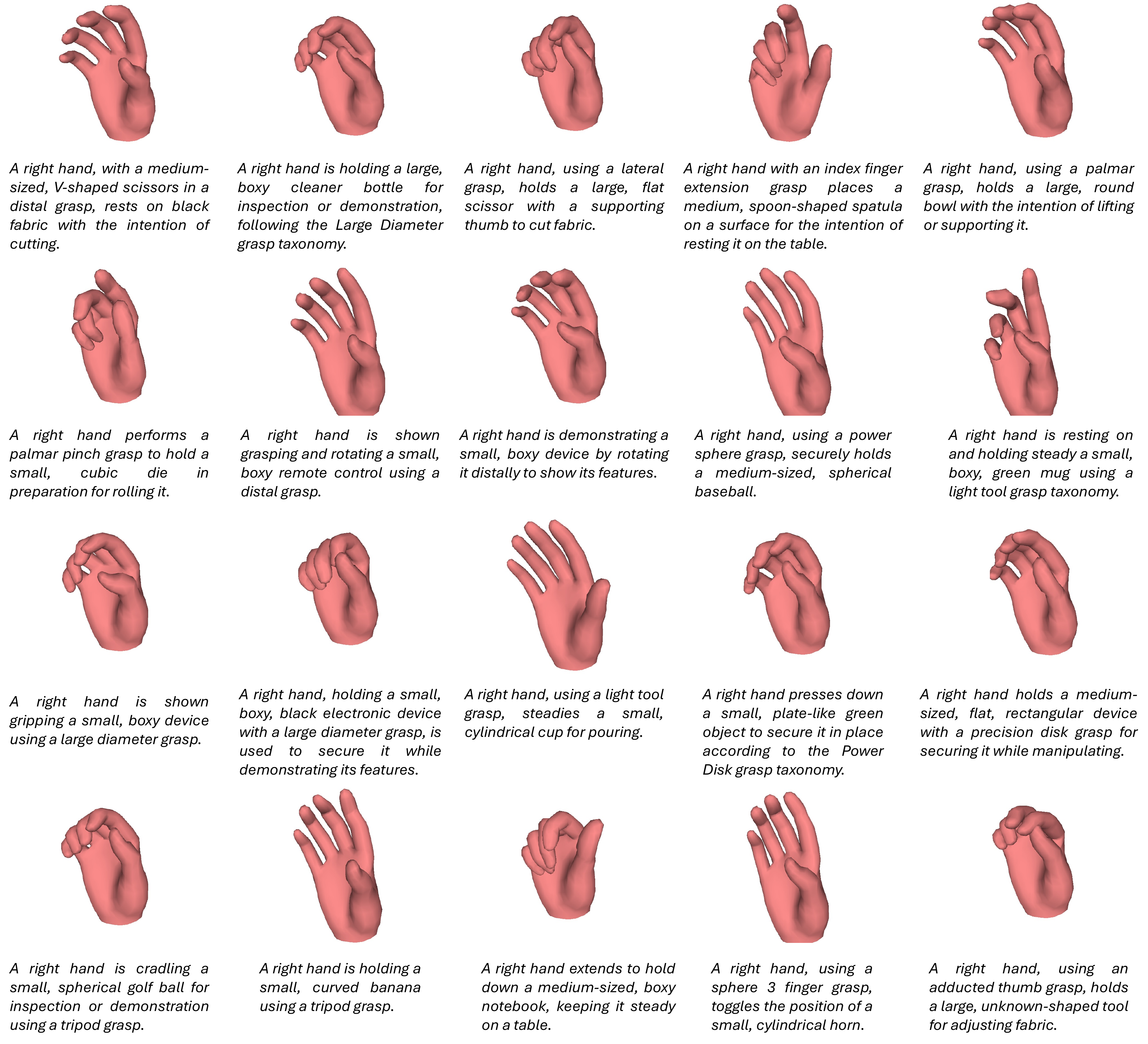}
    \caption{
    Sampled by our affordance-description conditioned prior.
    }
    \label{fig:qualitative_sample}
\end{figure*}

\customparagraph{Failure cases}
We finally inspect cases where our prior offers limited benefit (\cref{fig:hint_fail}).
For interactions far from the training distribution, such as unusual grasps or rare object affordances in in-the-wild HInt images, affordance descriptions can become less reliable, leading to weaker refinement over the initial WiLoR estimate.
Broadening the affordance vocabulary and grounding it in explicit 3D object geometry may help close this gap, as discussed in the limitations below.
\customparagraph{Text-conditional generation}
In Fig.~\ref{fig:qualitative_sample}, we provide the qualitative results of our diffusion-based pose sampling from the text input.
Our diffusion prior generates hand poses semantically aligned with affordance-aware descriptions, with high sample diversity.

\end{document}